\documentclass[10pt,twocolumn,letterpaper]{article}

\usepackage{iccv}
\usepackage{times}
\usepackage{epsfig}
\usepackage{graphicx}
\usepackage{amsmath}
\usepackage{amssymb}
\usepackage{algorithm} 
\usepackage{algpseudocode} 
\usepackage{float}

\restylefloat{algorithm}
\algnewcommand{\LineComment}[1]{\Statex \(\triangleright\) #1}
\renewenvironment{quote}{
  \list{}{
    \leftmargin0.3cm  
    \rightmargin\leftmargin
  }
  \item\relax
}
{\endlist}

\usepackage[title]{appendix}
\numberwithin{figure}{section}

\usepackage[breaklinks=true,bookmarks=false]{hyperref}

\iccvfinalcopy 

\def\iccvPaperID{5744} 

\ificcvfinal\fi

\begin{document}

\title{Posterior Sampling for Image Restoration using Explicit Patch Priors}

\author{{Roy Friedman \qquad Yair Weiss} \\
School of Computer Science and Engineering\\
The Hebrew University of Jerusalem\\
Jerusalem, Israel\\
{\tt\small \{roy.friedman,yair.weiss\}@mail.huji.ac.il}
}

\maketitle
\ificcvfinal\thispagestyle{empty}\fi

\begin{abstract}
   Almost all existing methods for image restoration are based on optimizing the mean squared error (MSE), even though it is known that the best estimate in terms of MSE may yield a highly atypical image due to the fact that there are many plausible restorations for a given noisy image. In this paper, we show how to combine explicit priors on {\em patches} of natural images in order to  sample from the posterior probability of a {\em full image} given a degraded image. We prove that our algorithm generates correct samples from the distribution $p(x|y) \propto \exp(-E(x|y))$ where $E(x|y)$ is the cost function minimized in previous patch-based approaches that compute a single restoration. Unlike  previous approaches that computed a single restoration using MAP or MMSE, our method makes explicit the uncertainty in the restored images and guarantees that all patches in the restored  images will be typical given the patch prior. Unlike previous approaches that used implicit priors on fixed-size images, our approach can be used with images of any size.  Our experimental results show that posterior sampling using patch priors yields images of high perceptual quality and high PSNR on a range of challenging image restoration problems. 
\end{abstract}


\begin{figure}
\begin{center}
\includegraphics[width=0.9\linewidth]{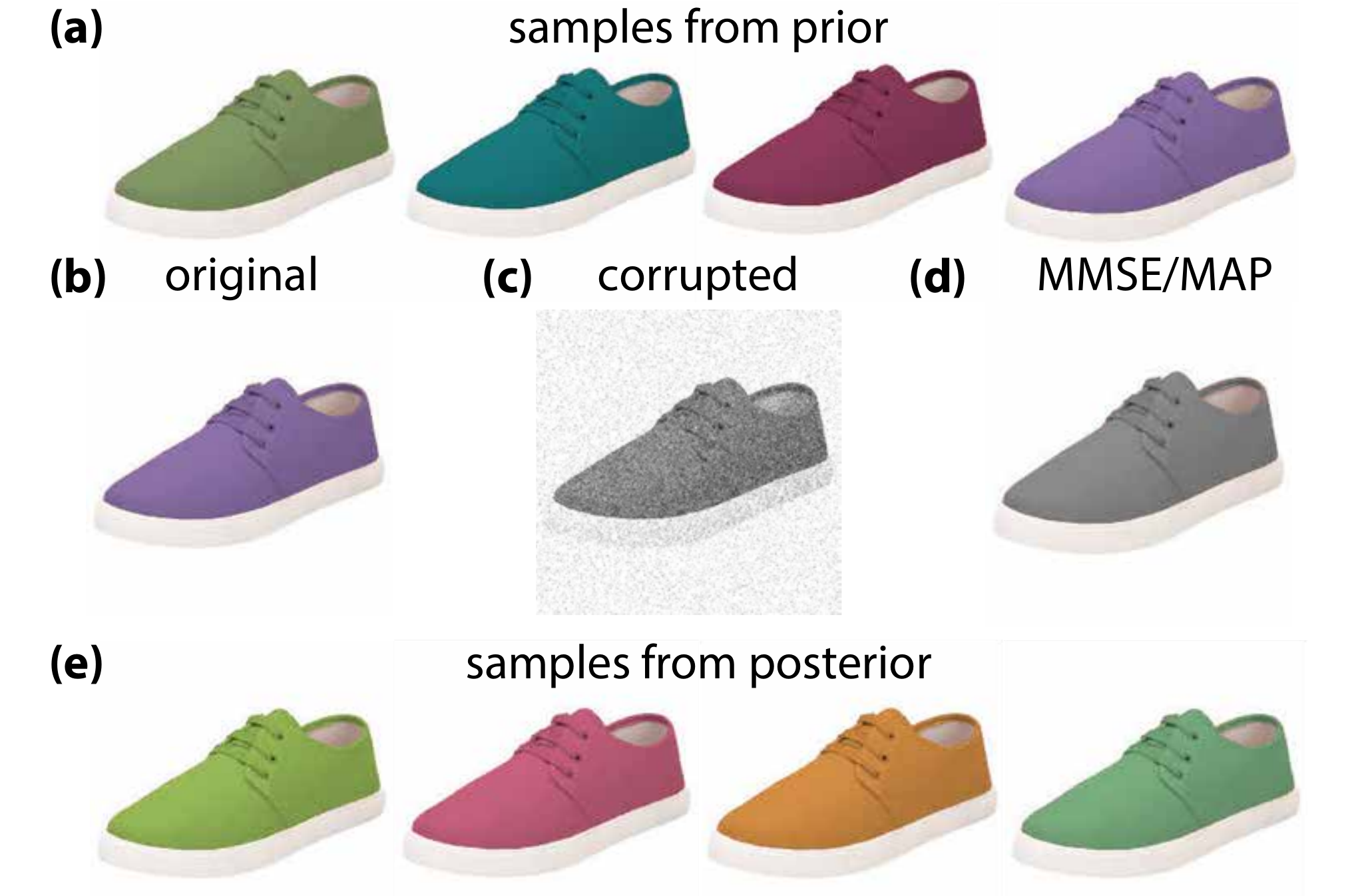} \\
\end{center}
   \caption{Posterior sampling yields images of high perceptual quality while the minimum mean squared error (MMSE) estimate does not. \textbf{(a)} Images are sampled from a Gaussian prior $p(x)$. When one of these images is converted to gray scale and corrupted with noise as in \textbf{(c)}, the MMSE estimate \textbf{(d)} averages over all possible colorizations and yields a gray shoe. In contrast, samples from the posterior \textbf{(e)} make explicit the possible different colorizations. In this paper, we seek to compute posterior samples for images by using explicit patch priors.}
   \label{fig:MMSE-posterior}
\end{figure}

\section{Introduction}
The task of image restoration includes many longstanding problems in computer vision such as denoising, deblurring, inpainting, demosaicing and super-resolution. All these problems can be phrased as recovering a ``clean image'' $x$ given a ``distorted image'' $y$. These problems are invariably ill-posed: for a given distorted image $y$, there are often an infinite number of clean images $x$ that could have given rise to $y$. 

In the recent decade, there has been continuous improvement in performance on these tasks, when measured according to the mean-squared error (MSE) between the restored image and the ground truth image. This is true for prior based methods (e.g.~\cite{JiSudderth,SchmidtGR10,DPIR,ZoranW11,KSVD,DabovFKE08,MichaeliI14,DBLP:conf/cvpr/ShocherCI18,Tirer19}), in which a prior over clean images $p(x)$ is combined with a likelihood $p(y|x)$, as well as deep learning methods that are trained end-to-end (e.g.~\cite{DBLP:conf/cvpr/KaufmanF20,8237753,DBLP:journals/corr/abs-1806-02919,DBLP:journals/corr/abs-1810-12575}).

In recent years, however, the limitations of using the peak signal-to-noise ratio (PSNR) or related evaluation scores such as the structural similarity index measure~\cite{SSIM} (SSIM) have become increasingly clear. Often, methods that produce the highest PSNRs yield restored images of low perceptual quality: they are often more blurred and of lower contrast compared to the ground-truth images\footnote{We follow the terminology of recent literature in defining an image as of "high perceptual quality" if its statistics are similar to those of natural images (e.g~\cite{BlauM18,wang2018esrgan}).}. As pointed out by Blau and Michaeli~\cite{BlauM18}, this trade-off between high PSNR and perceptual quality is due to the inherent ill-posedness of the image restoration problem.  To illustrate the trade-off, consider the toy image restoration problem shown in Figure~\ref{fig:MMSE-posterior}. The images are sampled from a Gaussian distribution $p(x)$  which generates shoes of different colors. Assume that we know $p(x)$ exactly and are given a new distorted image $y$ obtained by converting from color to gray scale and adding noise. How should $x$ be restored from $y$? It is well known that the optimal algorithm in terms of mean squared error is the algorithm that returns the conditional mean of $x|y$ (\cite{kay}). In this example, since the distribution is Gaussian, this is equivalent to calculating the MAP solution. Figure~\ref{fig:MMSE-posterior}(c) shows the MAP (or MMSE solution).  While this predictor is optimal in terms of PSNR, the generated image is completely gray, lacking the color of the original images.


Rather than defining a deterministic algorithm that computes the reconstruction that is optimal in the MSE sense, we seek a method that can sample \emph{multiple plausible reconstructions} given the corrupted image (cf.~\cite{ExplorableSR}). As shown in Figure~\ref{fig:MMSE-posterior}(e), if we sample from $p(x|y)$ each of the 4 plausible reconstructions of the noisy image have high perceptual quality. Thus rather than a deterministic algorithm that computes the mean reconstruction, we seek algorithms that can sample plausible reconstructions.  Mathematically, this amounts to sampling from the posterior distribution $p(x|y)$ rather than finding the maximum of this distribution or calculating its mean.

The idea of using posterior samples in the context of image restoration goes back to~\cite{DBLP:conf/icip/Fieguth03} and is relatively easy to operationalize when the distribution is Gaussian (e.g.~\cite{DBLP:conf/icip/Fieguth03,cheng2019bayesian}). Still, natural images are not Gaussian and, to the best of our knowledge, no existing algorithm can perform posterior sampling with an explicit non-Gaussian prior for a general image restoration problem. Some approaches train a deep neural network (DNN) to generate samples for a {\em single, fixed} problem~\cite{CondFlow, SRFlow}. Other approaches use the prior implicit in a denoiser to approximately generate samples from a posterior for a fixed-size, small image~\cite{EBIR,AGEM}. See Section~\ref{sec:related-work} for a detailed discussion of related work. 

The challenge in applying the idea of posterior sampling to image restoration is double. First, we need to define a prior probability $p(x)$ over full images that can capture the tremendous variety of natural images. Second, we need to devise an algorithm that can provably sample from $p(x|y)$ given a noisy observation $y$. 

In this paper we address both of these challenges. Our fundamental insight is that while defining a prior probability over full images remains a difficult problem, significant progress has been achieved with modeling probabilities of small image patches~\cite{ZoranW11,KSVD,DBLP:series/civ/HyvarinenHH09,10.1007/978-3-030-58542-6_15}. In fact, for some distributions such as image textures, a uniform distribution over a fixed dictionary of patches captures the distribution remarkably well~\cite{DBLP:conf/siggraph/EfrosF01,Barnes:2009:PAR,dahl2011learning,6528301}. To leverage this insight, we show how to use explicit priors over patches of natural images in order to perform posterior sampling over arbitrarily sized images and arbitrary image degradations. We prove that our algorithm generates samples from $p(x|y) \propto \exp(-E(x|y))$ where 
$E(x|y)$ is the cost function minimized in previous patch-based approaches that compute a single restoration. Unlike  previous approaches that computed a single restoration using MAP or MMSE, our method makes explicit the uncertainty in the restored images and guarantees that all patches in the restored  images will be typical given the patch prior. Additionally, unlike methods based on implicit priors, our approach can be used with images of any size and gives an analytic form for $p(x)$. Our experimental results show that posterior sampling using patch priors yields images of high perceptual quality and high PSNR on a range of challenging image restoration problems.

\section{Algorithm}

\begin{figure}[h!]
\begin{center}
\includegraphics[width=\linewidth]{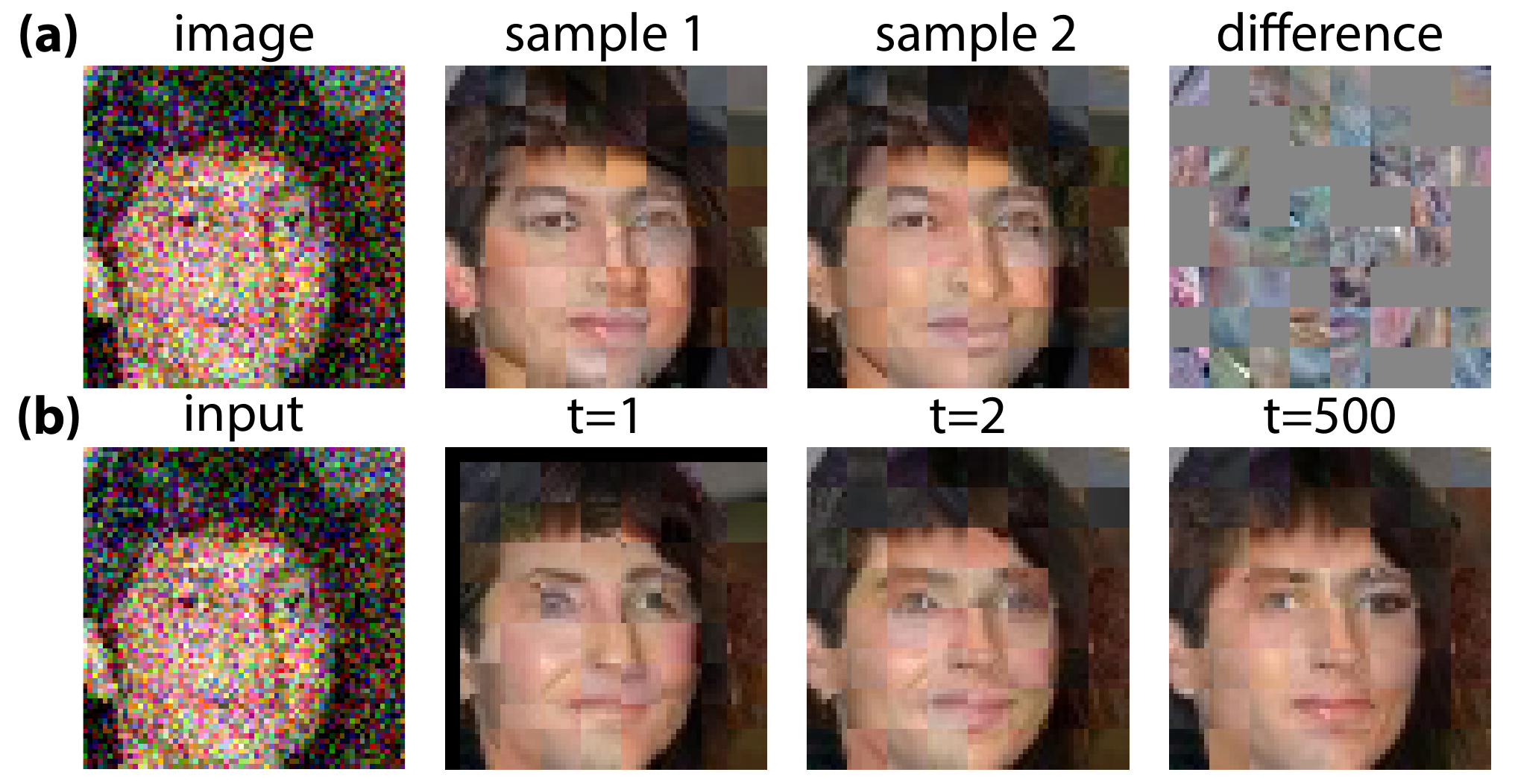}
\end{center}
    \caption[]{\textbf{(a)} Posterior samples with non-overlapping patches. Each patch is copied from one of the training images and is therefore realistic, and the samples show multiple plausible restorations, but the result is unsuitable for image restoration due to strong grid artifacts. \textbf{(b)} An illustration of our Gibbs sampling algorithm for posterior samples with overlapping patches using the same discrete dictionary prior as in (a). Each iteration consists of denoising using a different grid of non-overlapping patches. Over time, the grid artifacts are reduced and the algorithm converges to a clean image in which all patches in both grids are typical.}
    \label{fig:discrete}
\end{figure}



Denote by $x$ the unknown image, by $y=Hx+\eta$ the degraded image and by $P_i x$ the $i$th patch of the image. We assume that a patch model $p(P_i x)$ is available to us and that we can efficiently sample a patch from the posterior given a noisy observation $r$ of the same patch. We also assume that $\eta$ is Gaussian noise with known standard deviation $\sigma$ and that $H$ is given to us. Our algorithm can use {\em any} patch model, but for the sake of concreteness we first illustrate it in the case of a discrete dictionary of patches $p(P_i x=t)=\frac{1}{|\mathcal{D}|}$ if $t\in \mathcal{D}$ and $0$ otherwise. 
In the discrete case, the posterior probability is also discrete and we simply choose one of the patches $d_i$ in the dictionary with a probability proportional to $p(r|d_i)$. We now ask, given that we can sample patches from the posterior, how can we compute posterior samples over full images? 

\subsection{Exact Posterior Sampling with Non-Overlapping Patches}

Consider a set of {\em non-overlapping} patches that are organized in a grid $g$. We can consider this set of non-overlapping patches as a single image, which allows us to define a probability distribution over images by assuming all patches are independent. We denote this distribution by $p_g(x)=\prod_{i \in g} p(P_i x)$. Clearly sampling from this "prior" $p_g(x)$ is trivial: we only need to sample each patch independently. This is also true for posterior sampling given a noisy image that has been corrupted by independent noise at each pixel (i.e. $H$ is diagonal):  we sample each patch from $p(d_i|P_i y)$ independently and in this manner obtain a sample of the whole image from $p_g(x|y)$.

Figure~\ref{fig:discrete}(a) shows posterior samples from $p_g(x|y)$ for the noisy test image shown on the left, where we use non-overlapping $8 \times 8$ patches and the dictionary is constructed from a set of ~1300 training images (at each patch location, the dictionary consists of all training patches at the same location). By construction, each of the patches in the grid is realistic since it is sampled from one of the training images. Furthermore, the posterior samples capture the inherent uncertainty in the reconstruction: each sample gives a different plausible denoising of the image on the left. 
Despite these facts, this prior is clearly not suitable for image reconstruction. Both samples show noticeable grid artifacts.

\subsection{Exact Posterior Sampling with Overlapping Patches}

\begin{algorithm}[t] 
	\caption{Posterior Sampling Algorithm} 
	\hspace*{\algorithmicindent} \textbf{Input:} $H$, $y$, $\sigma^2$, $f(\cdot)$, $T$, $p_g(\cdot)$
	\begin{algorithmic}[1]
	    \State Initialize $x^{(0)}_1,\ldots,x^{(0)}_G$
	    \State Initialize $t$
		\For {$i=1,2,\ldots T$}
		    \For {$g=1,2,\ldots,G$}
		        \LineComment{\ \ \ \ \ \ \ \ \ Increase $\gamma,\beta$:}
        		\State $\beta,\gamma\leftarrow
        		f(i)$ 
        		\State $\bar{x} \leftarrow \frac{x^{(i)}_{g-1}+x^{(i)}_{g+1}}{2}$ 
        		\LineComment{\ \ \ \ \ \ \ \ \ Sample from a Gaussian:}
        		\State $x^{(i)}_g \sim e^{-\beta\|x_g-\bar{x}\|^2} e^{-\gamma\|x_g-t\|^2} e^{-\frac{1}{2 G \sigma^2}\|Hx_g-y\|^2}$
        		\LineComment{\ \ \ \ \ \ \ \ \ Sample non-overlapping patches:}
        		\State $t \sim e^{-\gamma\|x^{(i)}_g-t\|^2}p_g(t)$
        		
    		\EndFor
		\EndFor
		\State \textbf{return:} $x^{(T)}_G$
	\end{algorithmic} 
	\label{alg:posterior-sampling}
\end{algorithm}

While the above algorithm is simple to execute, the prior $p_g(x)$ is unrealistic; in natural images, neighboring patches are not independent. To this end, we now consider the case of overlapping patches. Even when the patches are overlapping, we can divide them into multiple sets of non-overlapping patches, each of which defines a grid. For example, consider all $8 \times 8$ patches in an image with a stride of $4$. These can be divided into four grids of which we will consider two: one grid contains patches whose top left corner has coordinates in $(1,9,17, \cdots)$ and the other contains patches whose top left corner has coordinates in $(5,13,21, \cdots)$.  In the previous section, we saw how to perform posterior sampling for each grid separately. How can we combine these samples?

We use the well-known method of half-quadratic splitting (e.g.~\cite{ZoranW11,DPIR}). We denote by $p_1(x)$ the distribution over images when considering only patches in the first grid (i.e. patches whose top left corner has coordinates in $(1,9,17, \cdots)$) and $p_2(x)$ the same for patches in the second grid (i.e. patches whose top left corner has coordinates in $(5,13,21, \cdots)$). We introduce two variables $x_1,x_2$ and define the following distribution over the two variables:

\begin{equation}
\begin{aligned}
p_\beta(x_1,x_2,y) =& p_1(x_1) \exp(-\frac{1}{4 \sigma^2} \|Hx_1 - y\|^2) \times \\
&p_2(x_2) \exp(-\frac{1}{4 \sigma^2} \|Hx_2 - y\|^2) \times \\
&\exp(-\beta\|x_1-x_2\|^2)
\end{aligned}
\end{equation}
As $\beta \rightarrow \infty$, $x_1$ and $x_2$ must equal each other. However, for a finite $\beta$ we can use the following Gibbs sampling algorithm:
 \vspace{-.85mm}
\begin{itemize}
\itemsep0pt
    \item Sample $x_1$ given $x_2,y$; this corresponds to posterior sampling with {\em nonoverlapping} patches from a ``noisy observation'' which is a weighted average of $x_2$ and $y$
    \item Sample $x_2$ given $x_1,y$; this corresponds to posterior sampling with {\em nonoverlapping} patches from  a ``noisy observation'' which is a weighted average of $x_1$ and $y$.
    \item Repeat
 \end{itemize}

Figure~\ref{fig:discrete}(b) shows posterior samples from this algorithm for the same prior and noisy image as in Figure~\ref{fig:discrete}(a) and a fixed $\beta=50$. In the first iterations, both $x_1$ and $x_2$ show strong grid artifacts but note that the artifacts appear in different locations in $x_1,x_2$ since they are based on different grids. After a few iterations the two images increasingly agree with each other and the grid artifacts are greatly reduced. If we continue the process, increasing $\beta$ throughout the process until $\beta \rightarrow \infty$, we will converge to a clean image where each patch in the image {\em in both grids} is sampled from the dictionary.

\begin{figure}[t]
\begin{center}
\includegraphics[width=\linewidth]{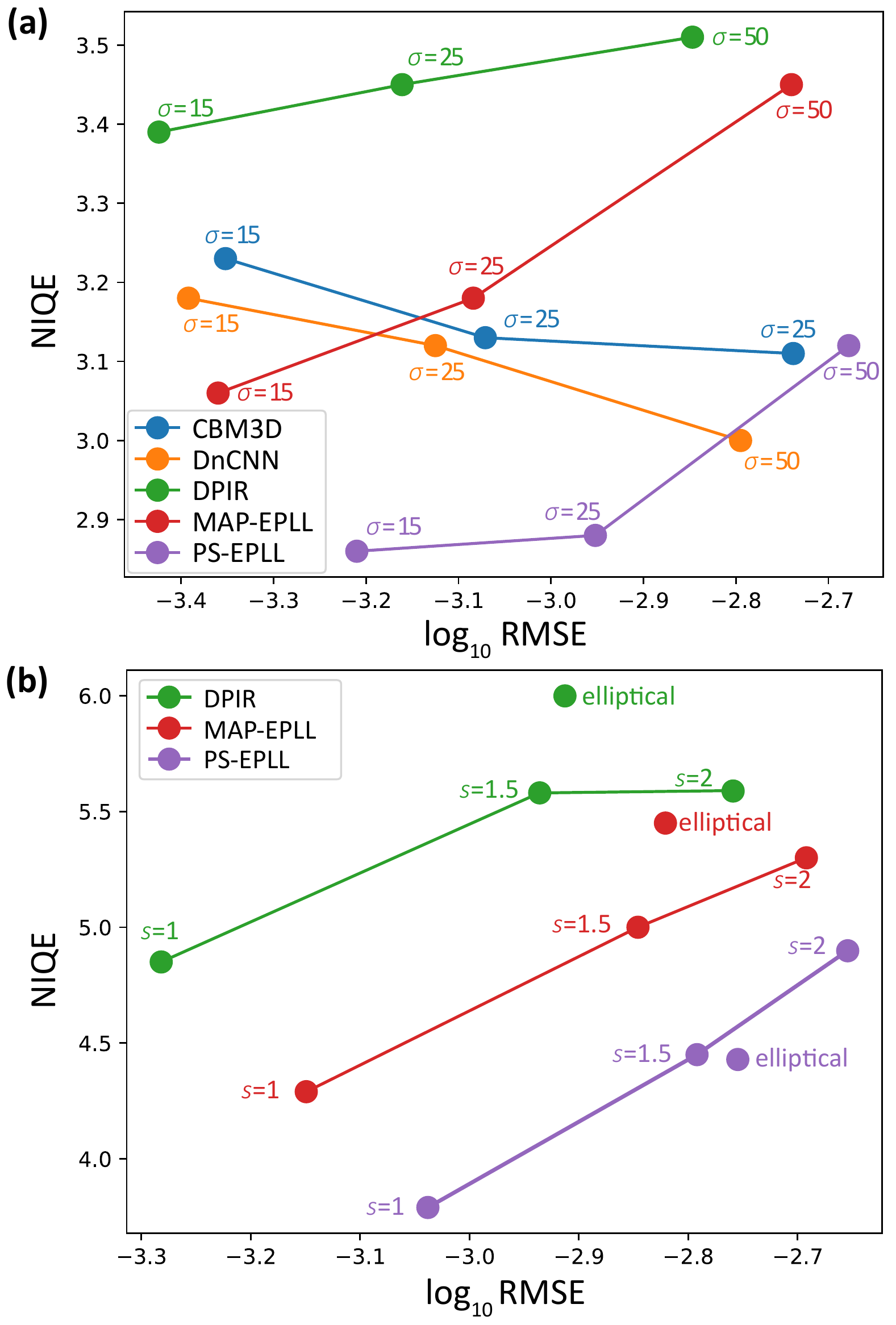}
\end{center}
  \caption{Perception-distortion graphs for \textbf{(a)} denoising and \textbf{(b)} deblurring using different models. Each curve in the graphs depicts a method, where each point depicts a particular parameter setting (i.e. standard deviation of noise or scale of blur kernel). In most cases, posterior sampling (purple points) has lower (better) NIQE compared to other approaches, at the cost of higher (worse) RMSE.}
\label{fig:perception-distortion}
\end{figure}


 
The algorithm can be generalized to any number of grids, $G$. To do so, we define $G$ variables $\{x_g\}$ and introduce a ``splitting'' term that requires them to be equal, as defined by the following distribution:
\begin{equation}
\begin{aligned}
    p_{\beta}\left(\{x_g\}_{g=1}^G \right.&\left.|y\right) = \frac{1}{Z} p_1(x_1) e^{-\frac{1}{2\sigma^2 G}\|Hx_1-y\|^2} \times \\
    & \prod_{g=2}^{G} p_g(x_g) e^{-\frac{1}{2\sigma^2 G}\|Hx_g-y\|^2}e^{-\beta\|x_g-x_{g-1}\|^2}
\end{aligned}
\label{eq:pbeta}
\end{equation}
where $Z$ is the normalization term for the distribution and we require each grid to be equal only to the previous grid. At infinite $\beta$, this will require all of the grids to agree.  Again, we can perform Gibbs sampling on $x_g$ and obtain a sample from $p_\beta(\cdot)$, as in the two grid example.

When $H$ is not diagonal (e.g. deblurring), then sampling may be difficult even when using non-overlapping patches. This is because the patches, which are independent in the prior, may no longer be independent given the corrupted image $y$. To address general $H$s, we add an additional auxiliary variable $t$ and define a joint distribution over $x,t,y$ via: 
\begin{equation} 
p_\gamma(x,t,y) \propto p_g(t) \exp(-\frac{1}{2 \sigma^2} \|Hx - y\|^2) \exp(-\gamma\|x-t\|^2)
\end{equation}
Now we can perofrm Gibbs sampling and sample $p(x|t,y)$ as if we were denoising non-overlapping patches with $t$ as the noisy image, and sample $p(t|x,y)$ by sampling from a multidimensional Gaussian.

The full algorithm is given in Algorithm~\ref{alg:posterior-sampling}. Note that it consists  of two simple steps that are iterated: (1) sampling each patch in a grid independently and (2) sampling from a multidimensional Gaussian. When $H$ is diagonal (e.g. in the case of denoising), the algorithm can be simplified so that it only requires the non-overlapping patches sampling step at each iteration (see supplementary material for the exact algorithm). In the next section, we show that the iteration of these two simple steps corresponds to sampling from a posterior probability over {\em full images}.  

\section{Analysis}

\begin{figure*}[h!]
\begin{center}
\includegraphics[width=\linewidth]{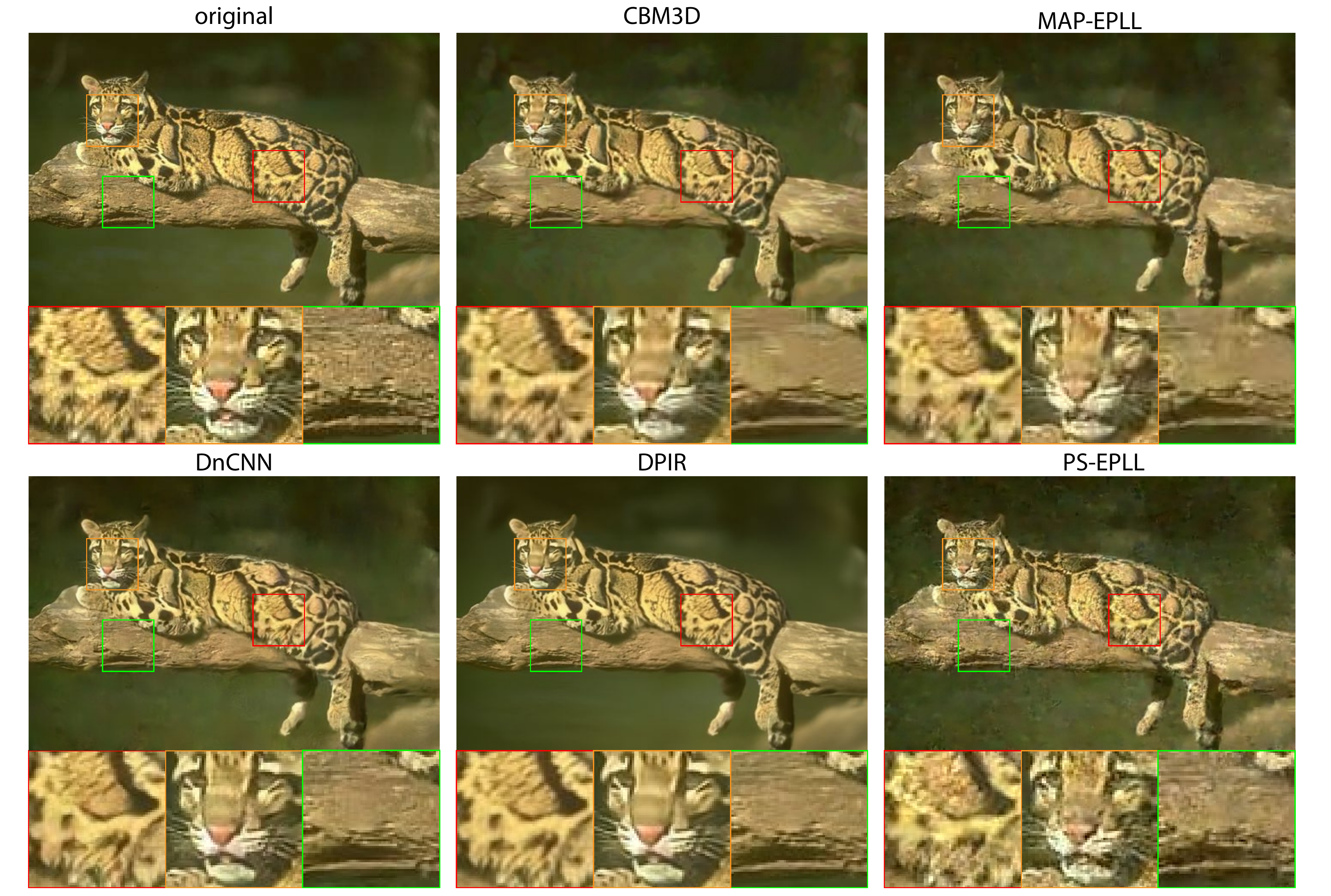}
\end{center}
  \caption{Visual comparisons of denoising methods for noise with $\sigma=50$ standard deviation; best viewed on a digital screen and zoomed in. Note how state-of-the-art denoisers truncate high frequencies while sampling from the posterior does not.}
\label{fig:denoise-BSDS}
\end{figure*}

\begin{figure}[h!]
\begin{center}
\includegraphics[width=\linewidth]{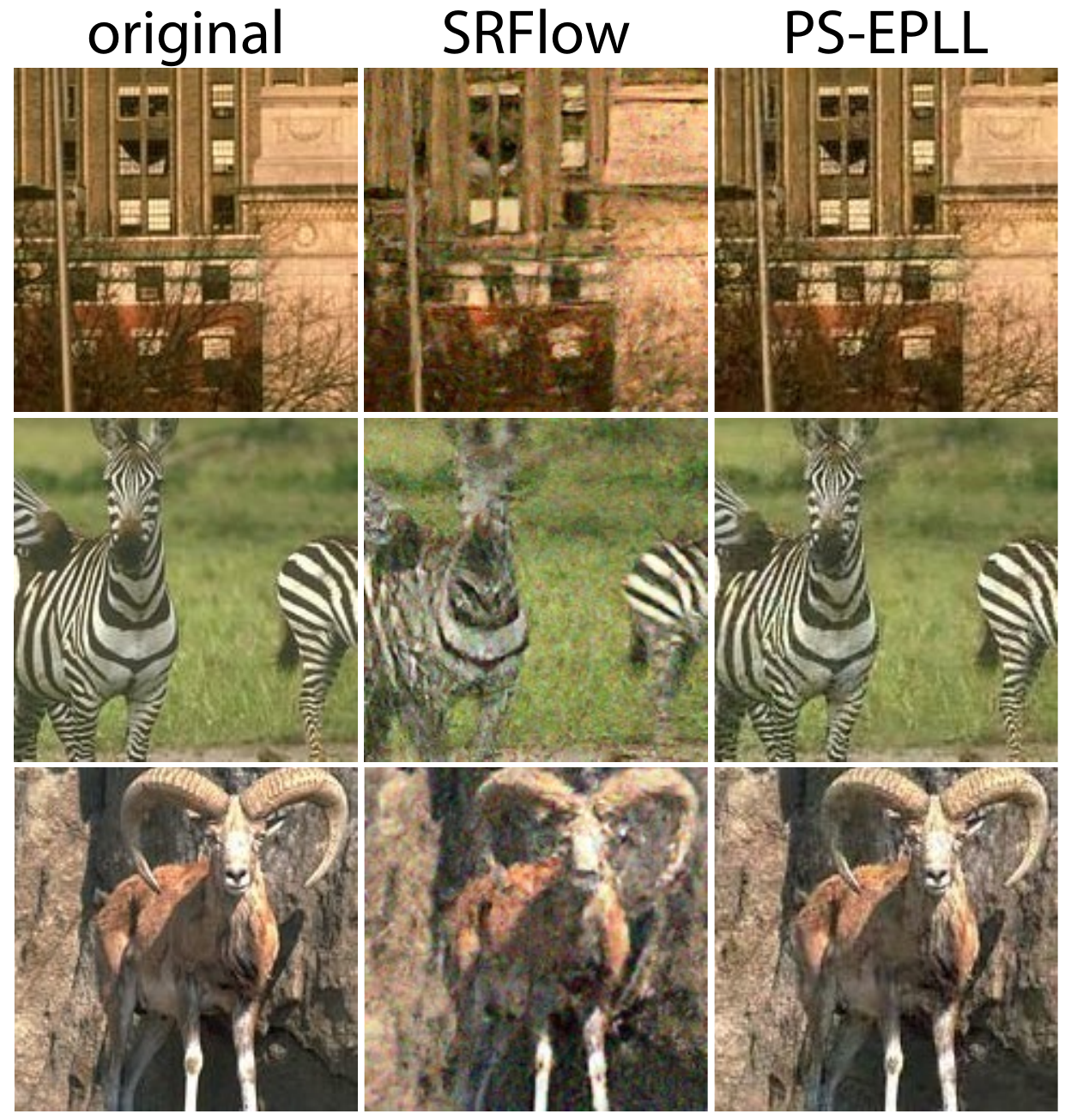}
\end{center}
  \caption{Visual comparisons of sampling solutions for denoising from SRFlow and with our method, with $\sigma=25$. SRFlow introduces strong artifacts and distortions, deviating quite a bit from the original images even for small noise levels, while our exact posterior sampling method does not.}
\label{fig:denoise-SRFlow}
\end{figure}


Our first result is for the case when the patch prior is uniform over a set of typical patches. This includes the case of a discrete dictionary over patches or the more expressive case of sparse coding dictionaries, which allows for any sparse combination  of dictionary elements to also be typical (e.g.~\cite{KSVD}).

{\bf Theorem 1:} If the patch prior is uniform over a set of typical patches, then as $T,\beta,\gamma \rightarrow \infty$ Algorithm~\ref{alg:posterior-sampling} converges to a posterior sample from $p(x|y)$ where the prior over images $p(x)$ is a uniform distribution on all images for which all patches are typical, assuming that the set of images for which all patches are typical is nonempty.

{\bf Proof:} This follows directly from equation~\ref{eq:pbeta} and the fact that we are performing Gibbs sampling. 

 Our second result shows that for any patch prior, samples from Algorithm~\ref{alg:posterior-sampling} are related to the Expected Patch Log Likelihood (EPLL) regularizer suggested by Zoran and Weiss in~\cite{ZoranW11}. In~\cite{ZoranW11} it was suggested to perform image restoration of full images given a patch prior $p(P_i x)$ by minimizing the following energy function:
\begin{equation}
    f_{EPLL}(x|y)= \lambda \|Hx -y\|^2 - EPLL(x)
\end{equation}
with $EPLL(x)=\sum_i \log p(P_i x)$
where $i$ ranges over all overlapping patches in the image. As the authors noted,  
\begin{quote}
    [This equation] has the familiar form of a likelihood term and a prior term, but note that $EPLL(x)$ is not the log probability of a full image. Since it sums over the log probabilities of all overlapping patches, it ``double counts'' the log probability. Rather it is the expected log likelihood of a randomly chosen patch in the image.
\end{quote}
 Nonetheless, this approach has been very successful for a wide range of image restoration problems and has been  used in many papers (e.g.~\cite{DBLP:conf/cvpr/PlotzR17,6528301,MultiScaleEPLL}). 

{\bf Theorem 2:} As  $T,\beta,\gamma \rightarrow \infty$  posterior samples from Algorithm 1 are samples from $p(x) \propto \exp(-f_{EPLL}(x|y))$.

{\bf Proof:} Again, this follows from equation~\ref{eq:pbeta} whose right hand side approaches $e^{-f_{EPLL}(x,y)}$ as $\gamma,\beta \rightarrow \infty$.


\begin{table*}[h]
\begin{center}
\resizebox{\linewidth}{!}{
\begin{tabular}{|r|c|c|c|c|c||c|c|c|c|c|c|}
\hline 
 & \multicolumn{5}{c||}{PSNR$\uparrow$} & \multicolumn{6}{c|}{NIQE$\downarrow$~ (clean~3.1$\pm$0.8)}\tabularnewline
\hline 
\hline 
$\sigma$ & CBM3D & DPIR & DnCNN & MAP-EPLL & PS-EPLL & noisy & CBM3D & DPIR & DnCNN & MAP-EPLL & PS-EPLL\tabularnewline
\hline 
15 & 33.5$\pm$2.0 & 34.2$\pm$2.1 & 33.9$\pm$2.0 & 33.6$\pm$1.9 & 32.1$\pm$1.5 &    5.2$\pm$0.9 & 3.2$\pm$0.6 & 3.4$\pm$0.8 & 3.2$\pm$0.7 & 3.1$\pm$0.6 & 2.9$\pm$0.6\tabularnewline
\hline 
25 & 30.7$\pm$2.2 & 31.6$\pm$2.4 & 31.3$\pm$2.3 & 30.8$\pm$2.2 & 29.5$\pm$1.9 &    10.8$\pm$2.0 & 3.1$\pm$0.5 & 3.5$\pm$0.8 & 3.1$\pm$0.6 & 3.2$\pm$0.5 & 2.9$\pm$0.6\tabularnewline
\hline 
50 & 27.4$\pm$2.5 & 28.5$\pm$2.6 & 28.0$\pm$2.5 & 27.4$\pm$2.4 & 26.8$\pm$2.25 &    18$\pm$3 & 3.1$\pm$0.5 & 3.5$\pm$0.8 & 3$\pm$0.4 & 3.4$\pm$0.5 & 3.1$\pm$0.6\tabularnewline
\hline 
\end{tabular}
}
\end{center}
\caption{Average PSNR (higher is better) and NIQE scores (lower is better) for denoising of images with noise levels 15, 25 and 50 on the BSD68 test images.}
\label{tab:BSDS-denoise-table}
\end{table*}

\begin{table*}[ht]
\begin{center}
\resizebox{\linewidth}{!}{
\begin{tabular}{|r|c|c|c||c|c|c|c|}
\hline 
 & \multicolumn{3}{c||}{PSNR$\uparrow$} & \multicolumn{4}{c|}{NIQE$\downarrow$~ (clean~3.1$\pm$0.8)}\tabularnewline
\hline 
\hline 
kernel & DPIR & MAP-EPLL & PS-EPLL & noisy & DPIR & MAP-EPLL & PS-EPLL\tabularnewline
\hline 
Isotropic, $s=1$ & 32.8$\pm$3.7 & 31.5$\pm$3.4 & 30.4$\pm$2.8 &    5.2$\pm$0.7 & 4.9$\pm$1.1 & 4.3$\pm$0.6 & 3.8$\pm$0.9\tabularnewline
\hline 
Isotropic, $s=1.5$ & 29.4$\pm$3.8 & 28.5$\pm$3.7 & 27.9$\pm$3.2 &    6.5$\pm$0.6 & 5.6$\pm$1.2 & 5.0$\pm$0.7 & 4.5$\pm$0.7\tabularnewline
\hline 
Isotropic, $s=2$ & 27.6$\pm$3.7 & 26.9$\pm$3.7 & 26.5$\pm$3.3 &    7.5$\pm$0.6 & 5.6$\pm$1.0 & 5.3$\pm$0.7 & 4.9$\pm$0.7\tabularnewline
\hline 
\hline
Elliptical kernel & 29.1$\pm$3.7 & 28.2$\pm$3.6 & 27.6$\pm$2.7 &    6.8$\pm$0.5 & 6.0$\pm$1.3 & 5.4$\pm$0.7 & 4.4$\pm$0.8\tabularnewline
\hline 
\end{tabular}
}
\end{center}
\caption{Average PSNR (higher is better) and NIQE scores (lower is better) for deblurring of BSD68 test images blurred with Gaussian kernels of different scales.}
\label{tab:BSDS-deblur-table}
\end{table*}

\section{Experiments}

\begin{figure*}[h]
\begin{center}
\includegraphics[width=\linewidth]{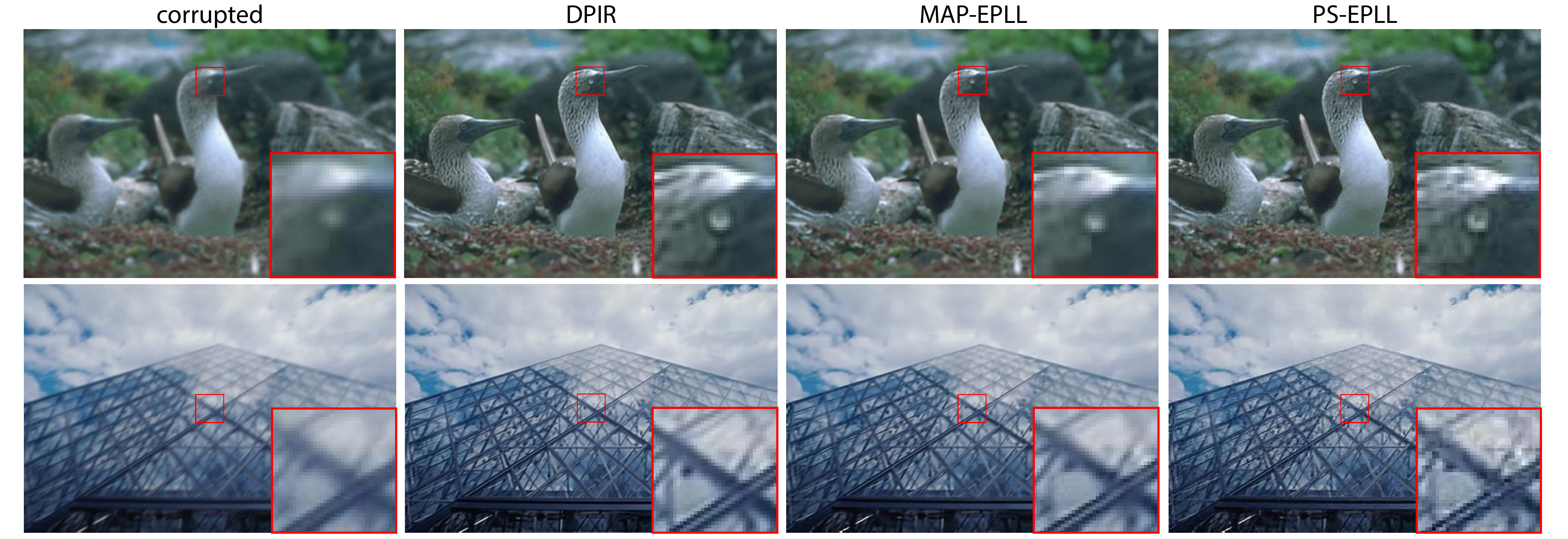}
\end{center}
  \caption{Visual comparisons of deblurring results; best viewed on a digital screen and zoomed in. The top row was blurred by a Gaussian kernel with scale $s=1.5$ while the bottom row was blurred by an elliptical (non-isotropic) Gaussian kernel.}
\label{fig:deblur-BSDS}
\end{figure*}

\begin{figure*}[h]
\begin{center}
\includegraphics[width=\linewidth]{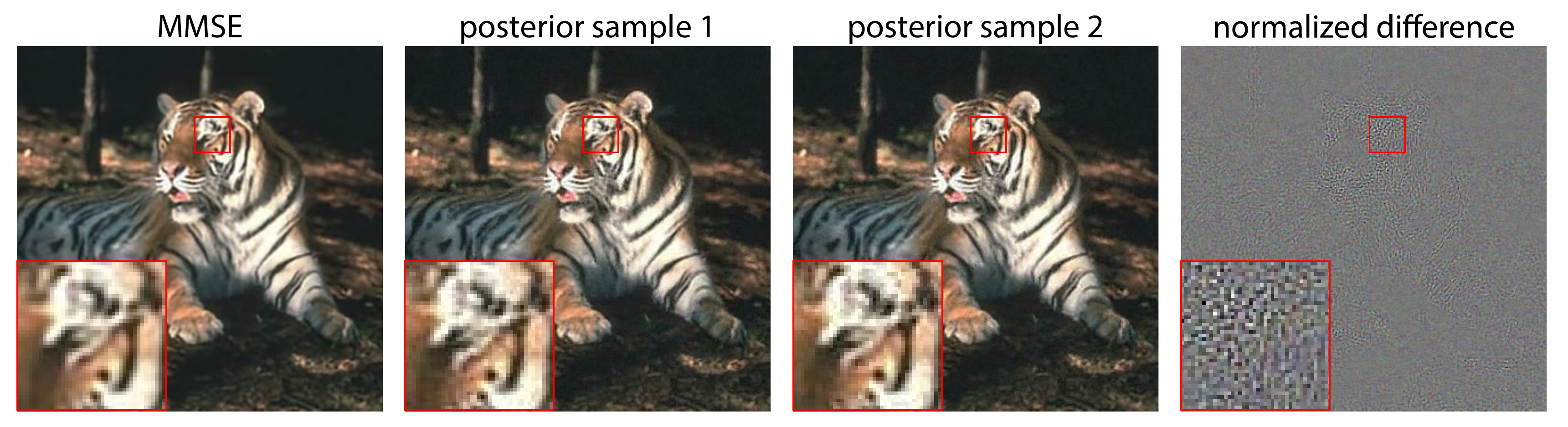}
\end{center}
  \caption{Comparison of the MMSE solution and posterior samples for image deblurring. The MMSE solution is blurrier than the posterior sample, even though its PSNR is higher. We also show the difference between two posterior samples, normalized to the range $[0,1]$.}
\label{fig:deblur-MMSE}
\end{figure*}

\begin{figure*}[h]
\begin{center}
\includegraphics[width=\linewidth]{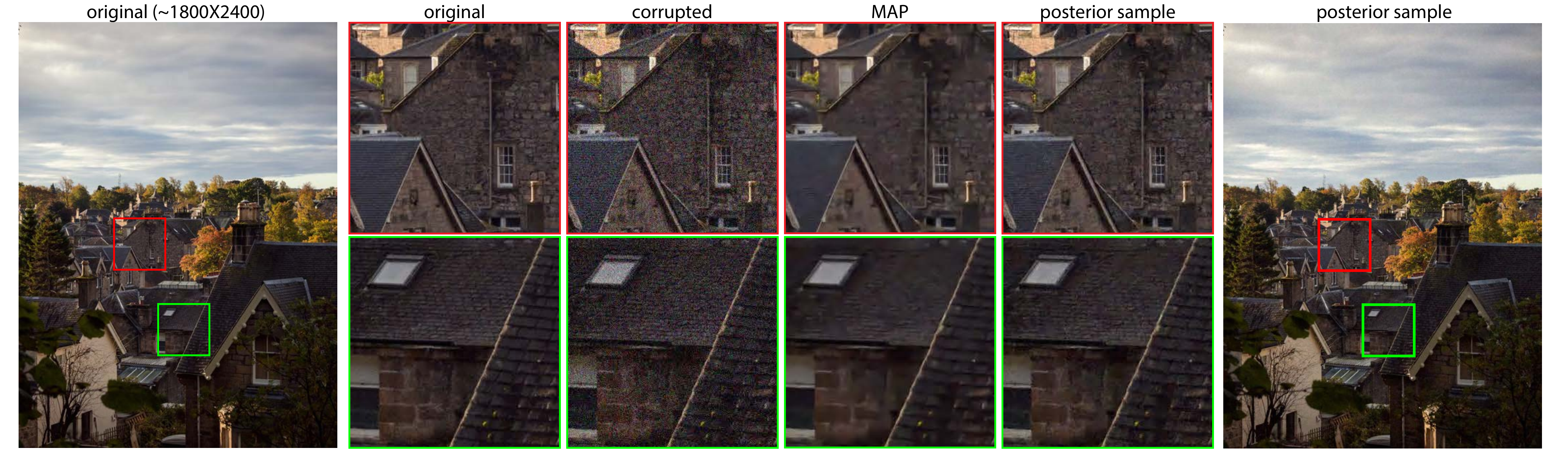}
\end{center}
  \caption{Denoising of a large, high resolution via posterior sampling. Particularly, note how the tiles on the roofs are retained in the posterior sample, but not in the MAP solution. Best viewed digitally while zoomed in.}
\label{fig:HR-denoise}
\end{figure*}

While our method can utilize any prior over patches, the experiments reported here use a GMM prior over image patches of size $8\times8\times3$ with 200 clusters. The GMM was trained exclusively on ~5 million patches of the 200 train images in the BSD300 dataset~\cite{BSDS}. See supplementary material for further information regarding the prior.

While our algorithm was originally devised to generate posterior samples, by replacing the sampling operations in lines 7 and 8 in Algorithm~\ref{alg:posterior-sampling} with a maximization operation we can obtain an approximate MAP algorithm for $p(x|y) \propto e^{-f_{EPLL}(x)}$, i.e. an approximate minimizer for $f_{EPLL}(x|y)$.  Throughout the experiments, we compare the posterior sampling version (indicated by PS-EPLL) with the MAP solutions (indicated by MAP-EPLL). As we show in the supplementary material, the MAP-EPLL algorithm gives almost identical performance to the EPLL algorithm in~\cite{ZoranW11} when using the same GMM prior.

For all of the image denoising and deblurring results, we used 100 iterations of our algorithm with 32 grids with a fixed update schedule for $\beta,\gamma$. The typical runtime for both of these tasks on images from the BSD68 test set was around 15 seconds using GPU computations, and around 45 seconds using the CPU. These runtimes can be further reduced by using less iterations on less grids or by making use of parallel computations over grids. Since the denoising problem involves a diagonal $H$, we use the simpler algorithm for denoising and the full Algorithm~\ref{alg:posterior-sampling} for deblurring.


To evaluate the perceptual quality of our samples, we used the well known NIQE~\cite{NIQE} score which measures the deviation between the local statistics in an image and the expected statistics in natural images. By evaluating the PSNR together with the NIQE score, we have an indication of the amount of distortion in the restorations while also measuring how dissimilar the statistics are to those of natural images.

{\bf Image Denoising:} In our first set of experiments, we performed image denoising on the BSD68 test set. We compare ourselves to a well known classical approach (CBM3D~\cite{CBM3D}), a state-of-the-art DNN denoiser that can handle many noise levels (DPIR~\cite{DPIR}) and another CNN method that was trained separately for each noise level (DnCNN~\cite{DnCNN}). 

As can be seen in Figure~\ref{fig:perception-distortion}(a) and Table~\ref{tab:BSDS-denoise-table}, our method for posterior sampling achieves lower NIQE (i.e. higher perceptual quality) at low noise levels, albeit at the cost of lower PSNRs. Furthermore, as can be seen in the perception-distortion plot in Figure~\ref{fig:perception-distortion}(a), our method does not have higher RMSE and NIQE, at the same time, compared to other methods in low noise levels. In the terms described by Blau and Michaeli~\cite{BlauM18}, our method is not dominated by any other (for low noise levels). Qualitatively, the generated posterior samples typically retain more textures and are sharper than other methods, as can be seen in Figure~\ref{fig:denoise-BSDS}. While the posterior samples do contain more high-frequency details (such as textures), they also incur visual artifacts for higher noise levels. We attribute these artifacts to the EPLL approach, which is defined only on small patches (i.e. $8\times8$). These artifacts are commonly seen in EPLL-based methods and have been shown to vanish when using a multi-scale approach in~\cite{MultiScaleEPLL}, which can be incorporated in our method as well.

As there aren't many methods that stochastically denoise (general subject) images, we also compare ourselves to SRFlow~\cite{SRFlow}, originally devised to create stochastic SR solutions using normalizing flows. SRFlow can also be used to denoise images stochastically, as suggested in~\cite{SRFlow}, by first down-sampling the corrupted input by a factor of 4 (using a bicubic kernel), essentially removing the noise from the image, before super resolving the image back to its original dimensions. As shown in Figure~\ref{fig:denoise-SRFlow}, sampling conditionally from SRFlow results in heavy artifacts and distortions in the image, even for the relatively low noise level of $\sigma=25$. Consistent with these visual comparisons, the observed PSNR of the SRFlow conditional samples was around 8 dB lower than the same for our conditional samples, at $\sigma=25$. Additional comparisons are shown in the supplementary material.

{\bf Deblurring:} Our second set of experiments was non-blind image deblurring. Once again, we used the BSD68 test set, corrupted by isotropic and non-isotropic Gaussian kernels. We used 3 isotropic kernels with different scales $s=[1,1.5,2]$ and an elliptical Gaussian kernel with $\sigma_x=1.5$, $\sigma_y=1$ and a correlation of $\rho=0.75$. After blurring the images, Gaussian noise with $\sigma=2.5$ standard deviation was added to the images. We again compared our method to DPIR~\cite{DPIR}, as it is a plug-and-play based model. Our posterior sampling method achieves better NIQE scores for all blur kernels, as can be seen in Figure~\ref{fig:perception-distortion}(b) and Table~\ref{tab:BSDS-deblur-table}. The posterior samples are also visibly much sharper than other methods, as seen in Figure~\ref{fig:deblur-BSDS}. To gain more insight into these results, we computed an approximate MMSE solution by averaging over 20 posterior samples, as can be seen in Figure~\ref{fig:deblur-MMSE}. The discrepancies between the posterior samples is barely discernible as shown in Figure~\ref{fig:deblur-MMSE}, but aggregation of these small changes in the high-frequencies results in the blurrier MMSE. At the same time, the MMSE solutions achieved higher PSNRs than the posterior samples while the MAP solutions achieved higher log-likehood values, although both are blurrier that posterior samples.

{\bf HR Denoising:} As previously stated, our method allows us to sample images of any size. Figure~\ref{fig:HR-denoise} shows that our method can be applied to high resolution images (results for ~1800x2400 are shown here) without any modification (the same number of iterations and grids was used). 

More examples for all of the above tasks can be found in the supplementary material, as well as results for inpainting.

\section{Related Work \label{sec:related-work}}

{\bf Stochastic Super Resolution:} Recently, there has been a growing body of work dealing with the creation of multiple possible solutions for a given super resolution (SR) problem~\cite{PULSE, ExplorableSR, ExplorableSR2}. These often stochastic methods typically produce sharper images, with finer details, than their deterministic counterparts which can be attributed to the extreme ill-posedness of the SR problem. Furthermore, these methods are commonly generalized to other image restoration problems that truncate high frequencies, by first down-sampling the degraded image and then using the SR output as the solution. However, most of these methods rely on an external mechanism in order to ensure some consistency between the high resolution output and the low resolution input and full consistency is unlikely. While these methods are stochastic they do not sample from the posterior probability, unlike our approach.

{\bf Flow Models:} Models utilizing methods of normalizing flows~\cite{CondFlow, SRFlow} are a promising line of work and allow high flexibility while still being explicitly invertible. By training a normalizing flow model on conditional samples, as presented by~\cite{CondFlow}, they have been used to solve a diverse set of image restoration tasks. However, a new model must be trained anew for each problem, and the samples are not explicitly consistent with their inputs for tasks involving non-diagonal corruption matrices, $H$. In contrast, our algorithm utilizes a single patch prior which can be used to restore any form of image corruption without any additional training and guarantees consistency by the fact that it is a posterior sample.

{\bf Implicit Denoiser Priors:} Modern NNs trained to denoise images achieve impressive performance and are a powerful tool. Utilizing the prior implicitly defined by a single denoiser~\cite{AGEM, SID, EBIR} has shown promising results for various image restoration tasks on small images. Unfortunately, as the name suggests, it is not entirely clear what this implicit prior actually models, i.e. the prior $p(x)$ used by these methods has no explicit form. Moreover, the sampling process is based on Langevin dynamics and similar methods, which can be very slow and unstable. For this reason, these approaches are almost always applied to small images (e.g. $96 \times 96$ images for~\cite{AGEM}, $256 \times 256$ images for~\cite{SID}).  In contrast, our algorithm is based on efficient patch sampling and can be applied to arbitrarily sized images, while still using an explicit prior.

\section{Conclusion}
As was pointed out by Fieguth in~\cite{DBLP:conf/icip/Fieguth03} two decades ago: 
\begin{quote}
    [Estimates for restored images based on MAP or MMSE] are not a realistic version of the random field, and do not represent a typical or representative sample of the system being studied. Instead, what is often desired is that we find a random sample from the posterior distribution, a much more subtle and difficult problem than estimation.
\end{quote} 
The challenge in applying this idea to general, real images is the inherent difficulty of calculating and sampling from $p(x)$ where $x$ is a full image. In this paper, we have taken an alternative approach where we assume we have access to a good prior over local patches of natural images. This allows for posterior sampling using \emph{any} prior over patches, as long as posterior sampling from each patch is possible. Using a simple GMM prior over patches, we have shown that such posterior samples are sharper and retain more of the properties typical to natural image than methods specializing in the minimization of the MSE, such as the MAP and MMSE estimates. Consistent with past observations, these improvements in the perceptual quality come at the price of the PSNR, but even after this drop the PSNR is quite high. Further still, posterior samples allow us to make explicit the uncertainty in obtaining restorations of degraded images. 


\section*{Acknowledgements}
We thank the Gatsby Foundation for financial support.

{\small
\bibliographystyle{unsrt}
\bibliography{biblio}
}

\newpage

\begin{appendices}

\newcommand{\cmvn}[3]{\ensuremath{\mathcal{N}\left(#1\ |\ #2, #3\right)}}
\newcommand{\mvn}[2]{\ensuremath{\mathcal{N}\left(#1, #2\right)}}

\section{GMM as a Patch Prior}
As stated in the main text, we used a GMM prior over patches, i.e.:
\begin{equation}
    p(x^i)=\sum_k^K \pi_k \cmvn{x^i}{\mu_k}{\Sigma_k}
\end{equation}
where $x^i\stackrel{\Delta}{=}P_ix$ is the $i$-th patch of the image $x$, $K$ is the number of clusters, $\pi_k$ are the cluster probabilities, $\mu_k$ are the cluster means and $\Sigma_k$ are the cluster covariance matrices. 

\subsection{Sampling from a GMM}
As shown in Algorithm 1 from the main text, we mainly need to know how to sample from the posterior when denoising in order to sample from the posterior for a general image restoration task. To this end, let $y^i=x^i+\eta$ where $y^i$ is a corrupted patch, $x^i$ is the clean patch we want to find and $\eta$ is some zero-mean noise. For our purposes, we will assume $\eta\sim\mvn{0}{I\sigma^2}$ where $\sigma$ is known to us, however in general the noise need not be Gaussian.

To perform denoising, we will have to find the following conditional:
\begin{equation}
    p(x^i|y^i)=\frac{1}{Z}e^{-\frac{1}{2\sigma^2}\|x^i-y^i\|^2}p(x)
\end{equation}
where $Z$ is a normalization constant. The above is the product of a Gaussian with a GMM, which will be a GMM given by:

\begin{equation}\label{eq:patch-post}
    p(x^i|y^i)=\sum_k p(k|y^i)\cmvn{x^i}{\mu^{(k)}_{x^i|y^i}}{\Sigma^{(k)}_{x^i|y^i}}
\end{equation}

\begin{equation}\label{eq:patch-log-marginal}
    p(k|y^i)=\frac{\pi_k \cmvn{y^i}{\mu_k}{\Sigma_k+I\sigma^2}}{\sum_{k'}\pi_{k'} \cmvn{y^i}{\mu_{k'}}{\Sigma_{k'}+I\sigma^2}}
\end{equation}
\begin{equation}\label{eq:patch-post-mean}
    \mu^{(k)}_{x^i|y^i}=\Sigma^{(k)}_{x^i|y^i}\left(\Sigma_k^{-1}\mu_k+\frac{1}{\sigma^2}y^i\right)
\end{equation}
\begin{equation}\label{eq:patch-post-cov}
    \Sigma^{(k)}_{x^i|y^i}=\left(\Sigma_k^{-1} + \frac{1}{\sigma^2}I\right)^{-1}
\end{equation}

Sampling a patch from the posterior in this case can be done in two steps:
\begin{enumerate}
    \item Sample a cluster $k$ from $k|y^i$, given by Equation~\ref{eq:patch-log-marginal}
    \item Sample $x^i$ from a Gaussian with mean $\mu^{(k)}_{x^i|y^i}$ and covariance $\Sigma^{(k)}_{x^i|y^i}$, given by Equations~\ref{eq:patch-post-mean} and~\ref{eq:patch-post-cov} respectively
\end{enumerate}

\subsection{Quality of the Prior}

\begin{figure}[h!]
\begin{center}
\includegraphics[width=\linewidth]{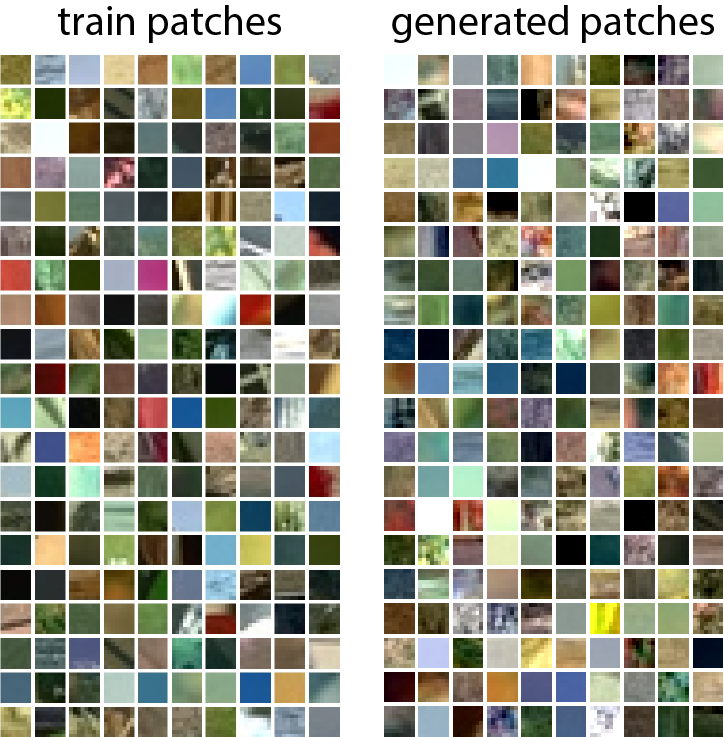}
\end{center}
    \caption{Examples of real patches (left) versus patches generated by our trained GMM (right).}
    \label{fig:prior-qual}
\end{figure}

We trained a GMM over 5 million patches from the train set of BSD300, using the expectation maximization (EM) algorithm, for 30 iterations. Figure~\ref{fig:prior-qual} shows samples drawn from the GMM (right) next to random patches from the train set (left).  

\section{Equivalence with EPLL}

\begin{table}[h]
\begin{center}
\resizebox{\linewidth}{!}{
\begin{tabular}{|r|c|c|}
\hline
$\sigma$ & EPLL~\cite{ZoranW11} & MAP-EPLL (ours) \tabularnewline
\hline 
\hline
15 & 31.21 & 31.29 \\
\hline
25 & 28.71 & 28.73 \\
\hline
50 & 25.72 & 25.66  \\
\hline 
\end{tabular}
}
\end{center}
\caption{Comparison of PSNR between our method and the original EPLL algorithm. As can be seen, both methods achieve nearly the same PSNR.}
\label{tab:EPLL-denoise-table}
\end{table}

While the MAP version of our algorithm find a local maximum of the EPLL energy term, it is not necessarily equivalent to the EPLL algorithm. To show that the performance of both methods is similar, we used the same GMM as was used in~\cite{ZoranW11} and compared the performance of both methods. The results of this experiment can be seen in Table~\ref{tab:EPLL-denoise-table} and show the performance is very similar.

\section{Denoising Algorithm}

\begin{algorithm} 
	\caption{Denoising Algorithm} 
	\hspace*{\algorithmicindent} \textbf{Input:} $H$, $y$, $\sigma^2$, $f(\cdot)$, $T$, $p_g(\cdot)$
	\begin{algorithmic}[1]
	    \State Initialize $x^{(0)}_1,\ldots,x^{(0)}_G$
	    \State Initialize $t$
		\For {$i=1,2,\ldots T$}
		    \For {$g=1,2,\ldots,G$}
                \LineComment{\ \ \ \ \ \ \ \ \ Increase $\beta$:}
        		\State $\beta\leftarrow
        		f(i)$ 
        		\State $\bar{x} \leftarrow \frac{x^{(i)}_{g-1}+x^{(i)}_{g+1}}{2}$ 
        		\LineComment{\ \ \ \ \ \ \ \ \ Sample independent patches using Eq.~\ref{eq:patch-post}:}
        		\State $x^{(i)}_g \sim e^{-\beta\|x_g-\bar{x}\|^2}e^{-\frac{1}{2 G \sigma^2}\|x_g-y\|^2}p_g(x_g)$
        		
    		\EndFor
		\EndFor
		\State \textbf{return:} $x^{(T)}_G$
	\end{algorithmic} 
	\label{alg:denoise}
\end{algorithm}

\begin{figure*}[h!]
\begin{center}
\includegraphics[width=\linewidth]{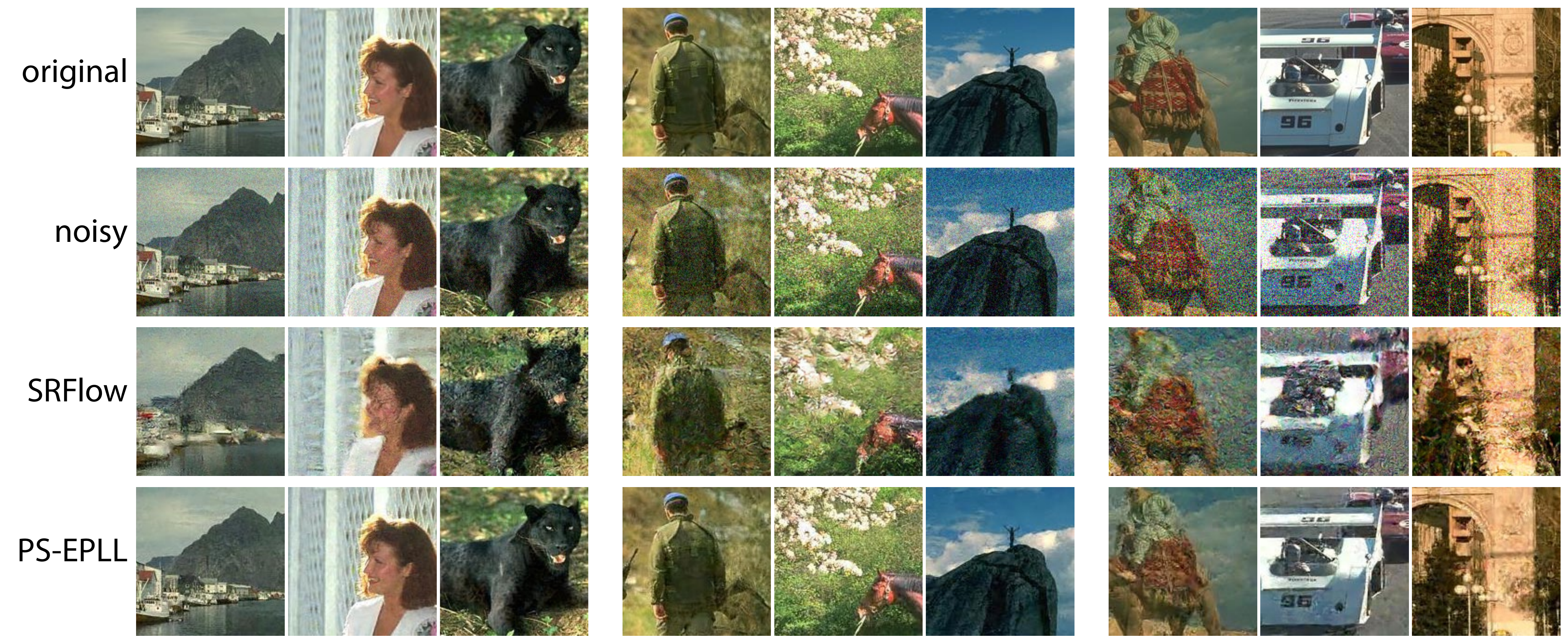}
\end{center}
    \caption{SRFlow with 8X downsampling. The left 3 rows are examples of denoisings with $\sigma=15$ noise, the middle 3 rows contain denoisings with $\sigma=25$ and the right 3 rows with $\sigma=50$.}
    \label{fig:srflow-8}
\end{figure*}

In the main body of text we presented the general posterior sampling algorithm, however a simplified version can be used for denoising. Algorithm~\ref{alg:denoise} depicts this simplified algorithm. In words, the second splitting term that was added in the original algorithm, separating the denoising and data steps, is no longer needed. This is due to the fact that in the denoising task, the corruption matrix $H$ is assumed to be the identity matrix.

For inpainting, a modified version of Algorithm~\ref{alg:denoise} was used, where instead of spherical noise $\eta$ a diagonal covariance was assumed, i.e. $\eta\sim\mvn{0}{D}$ where $D$ is a diagonal matrix. In this case, the only change is that every account of $I\sigma^2$ in Equations~\ref{eq:patch-log-marginal}--\ref{eq:patch-post-cov} needs to be exchanged with the matrix $D$, and every appearance of $I\frac{1}{\sigma^2}$ needs to be changed with $D^{-1}$.

\section{Choice of Hyperparameters}
For all of our experiments, the update schedules for $\beta$ and $\gamma$ were of the form:
\begin{align}
    f_\beta(i)=&\left(1+\left(\frac{i}{\tau_\beta}\right)^{d_\beta}\right)\cdot a_\beta \\
    f_\gamma(i)=&\left(1+\left(\frac{i}{\tau_\gamma}\right)^{d_\gamma}\right)\cdot a_\gamma
\end{align}
Further, for some of the tasks an additional step with a much larger $\beta$ (100$\times$ the maximal value) was used - we found that this additional step smoothed out any grid artifacts which may have remained after the sampling procedure.

In each of the tasks (subtly) different schedules were used. For all of them, $d_\beta=2.2$ was the same. The rest of the parameter settings were:
\begin{enumerate}
    \item $\tau_\beta$: for denoising and deblurring it was 18, for inpainting 6 
    \item $a_\beta$: for denoising $\frac{1}{\sigma^2}$, for inpainting 2, for deblurring 10
    \item For deblurring there were the additional parameters: $\tau_\gamma=1$, $d_\gamma=0.65$, $a_\gamma=0.1$
\end{enumerate}

We found that the ordering of the grids in the Gibbs sampling procedure and the schedule are dependent on each other. In our experiments, the order the grids were sampled in was from the first to the last and then from the last to the first, i.e. $1,2,\ldots,G-1,G,G-1,\ldots,2,1,2,\ldots$. For the first pass over the grids, only the previous grid in the chain was considered, as an initialization of the grids.

Finally, for the MAP solution we also returned the mean over all of the grids instead of just the last grid in the chain. We found that this typically boosted the PSNR, but also the EPLL score of the result. In all of the results displayed, the MAP had a higher EPLL score than both the MMSE and the posterior samples, as we should expect from a MAP estimate. 

\section{SRFlow with 8X Downsampling}

In the main text, comparisons with SRFlow were given only with 4X downsampling. In this case, the downsampling may not completely remove traces of the noise before sampling again. Figure~\ref{fig:srflow-8} shows comparisons between SRFlow and our posterior sampling algorithm when the images are downsampled by a factor of 8 before reconstructions are sampled from SRFlow. As can be seen, SRFlow samples add significant distortions to the reconstructions while our posterior samples do not. 

\section{More Examples}
\begin{figure*}[h!]
\begin{center}
\includegraphics[width=\linewidth]{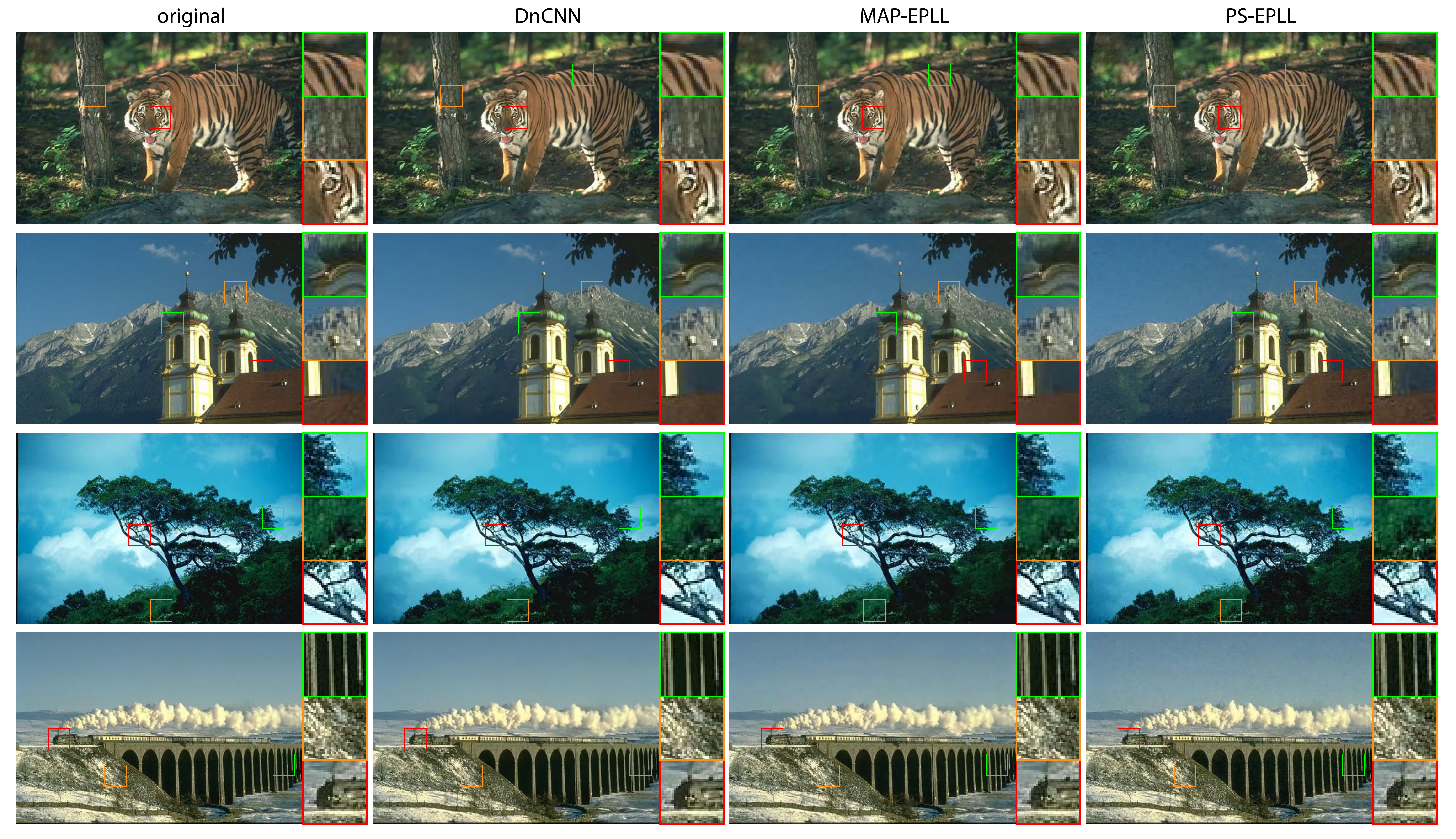}
\end{center}
    \caption{More examples of image denoising with $\sigma=15$ noise.}
    \label{fig:denoise-n15}
\end{figure*}

\begin{figure*}[h!]
\begin{center}
\includegraphics[width=\linewidth]{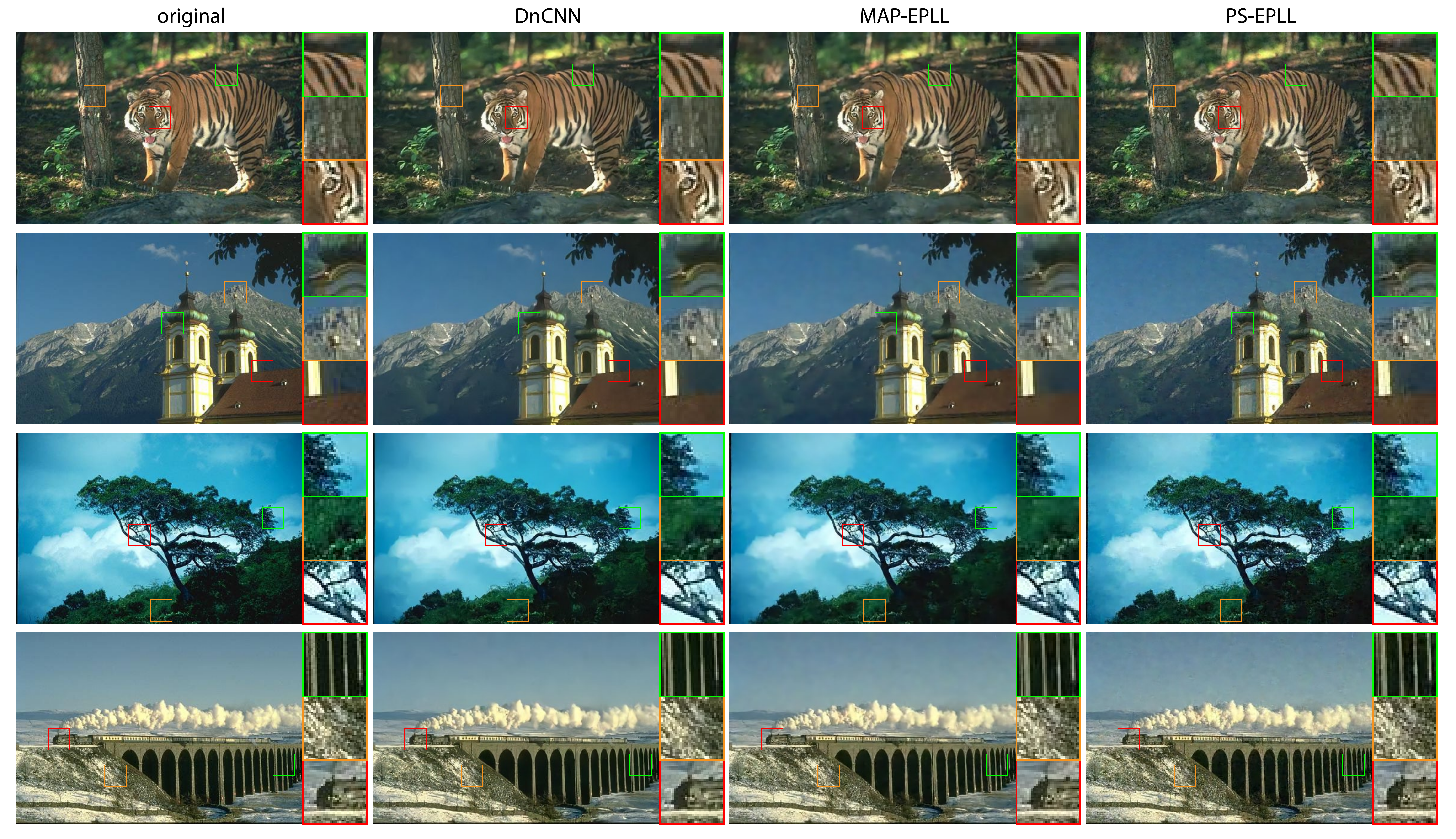}
\end{center}
    \caption{More examples of image denoising with $\sigma=25$ noise.}
    \label{fig:denoise-n25}
\end{figure*}

\begin{figure*}[h!]
\begin{center}
\includegraphics[width=\linewidth]{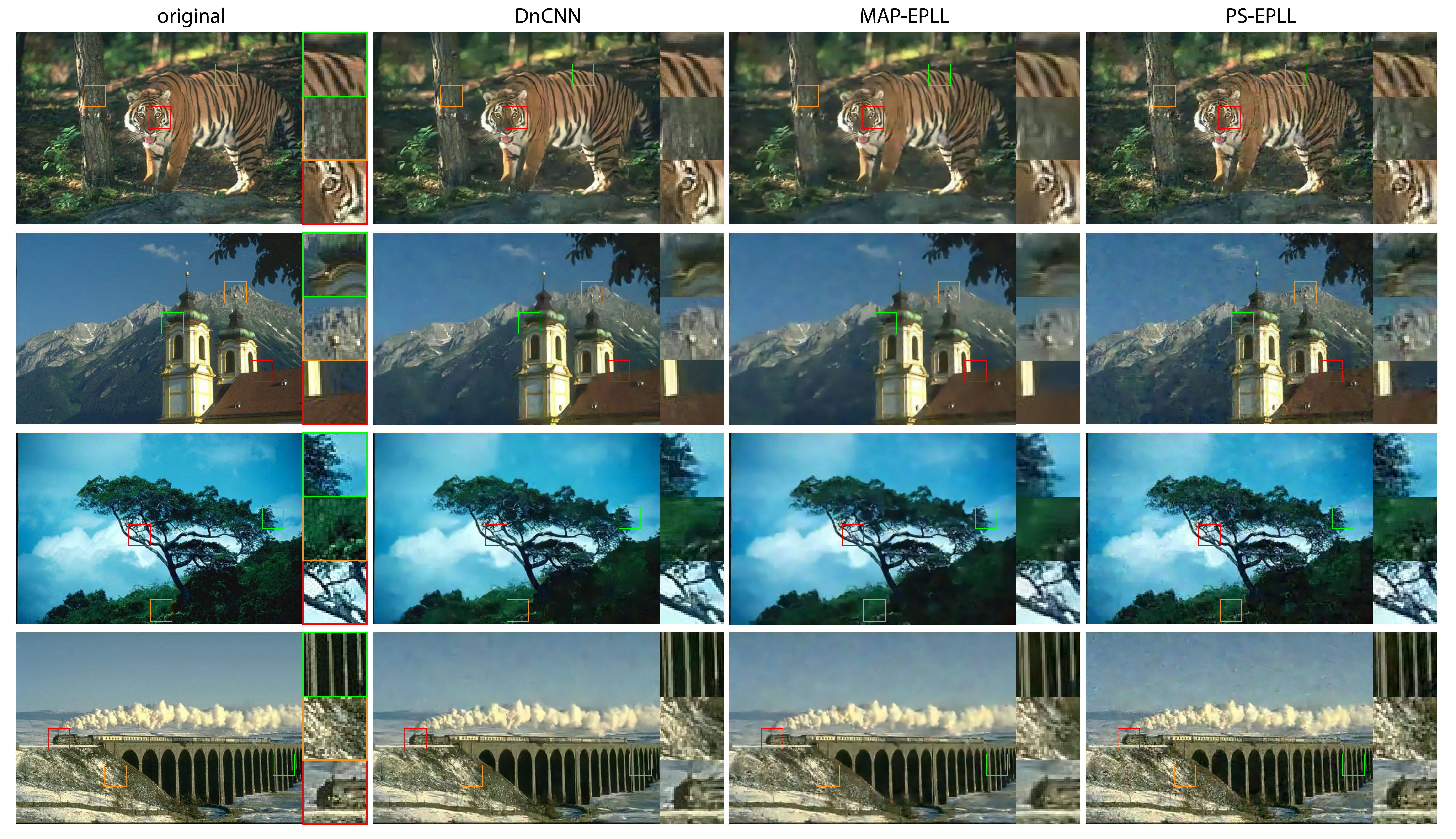}
\end{center}
    \caption{More examples of image denoising with $\sigma=50$ noise.}
    \label{fig:denoise-n50}
\end{figure*}

\begin{figure*}[h!]
\begin{center}
\includegraphics[width=\linewidth]{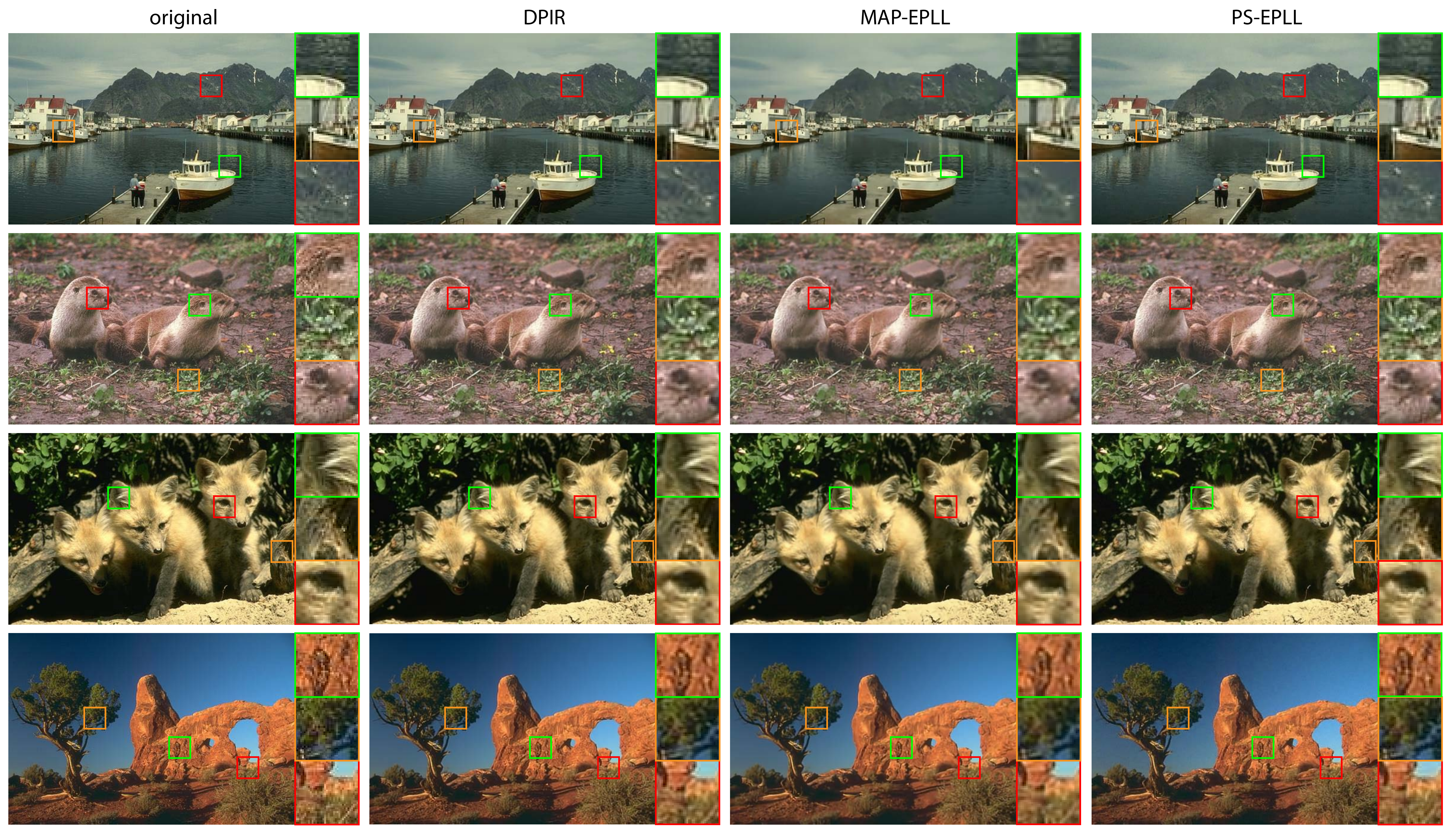}
\end{center}
    \caption{More examples of image deblurring for a Gaussian blur kernel with scale $s=1$}
    \label{fig:deblur-s1}
\end{figure*}

\begin{figure*}[h!]
\begin{center}
\includegraphics[width=\linewidth]{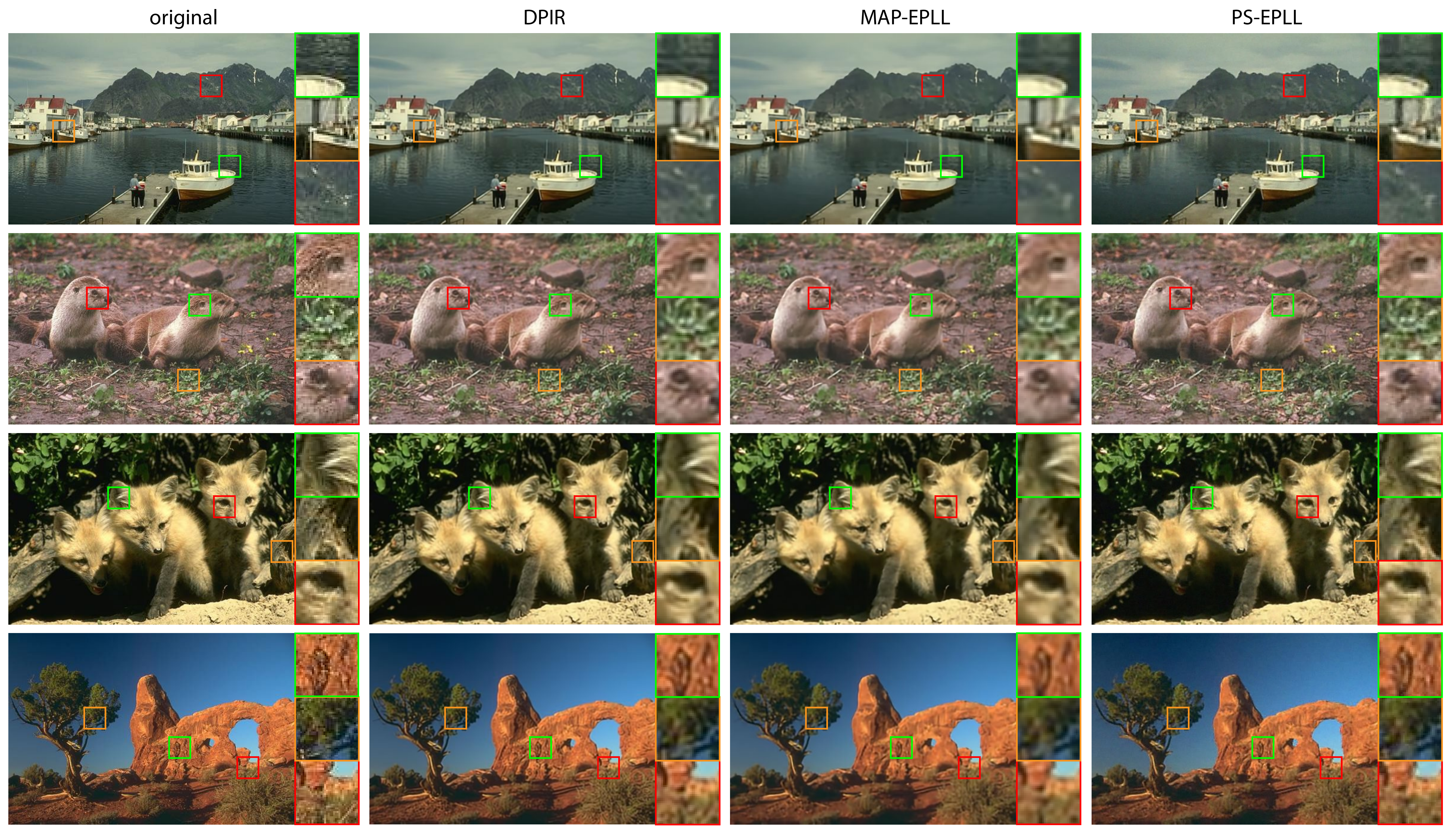}
\end{center}
    \caption{More examples of image deblurring for a Gaussian blur kernel with scale $s=1.5$}
    \label{fig:deblur-s15}
\end{figure*}

\begin{figure*}[h!]
\begin{center}
\includegraphics[width=\linewidth]{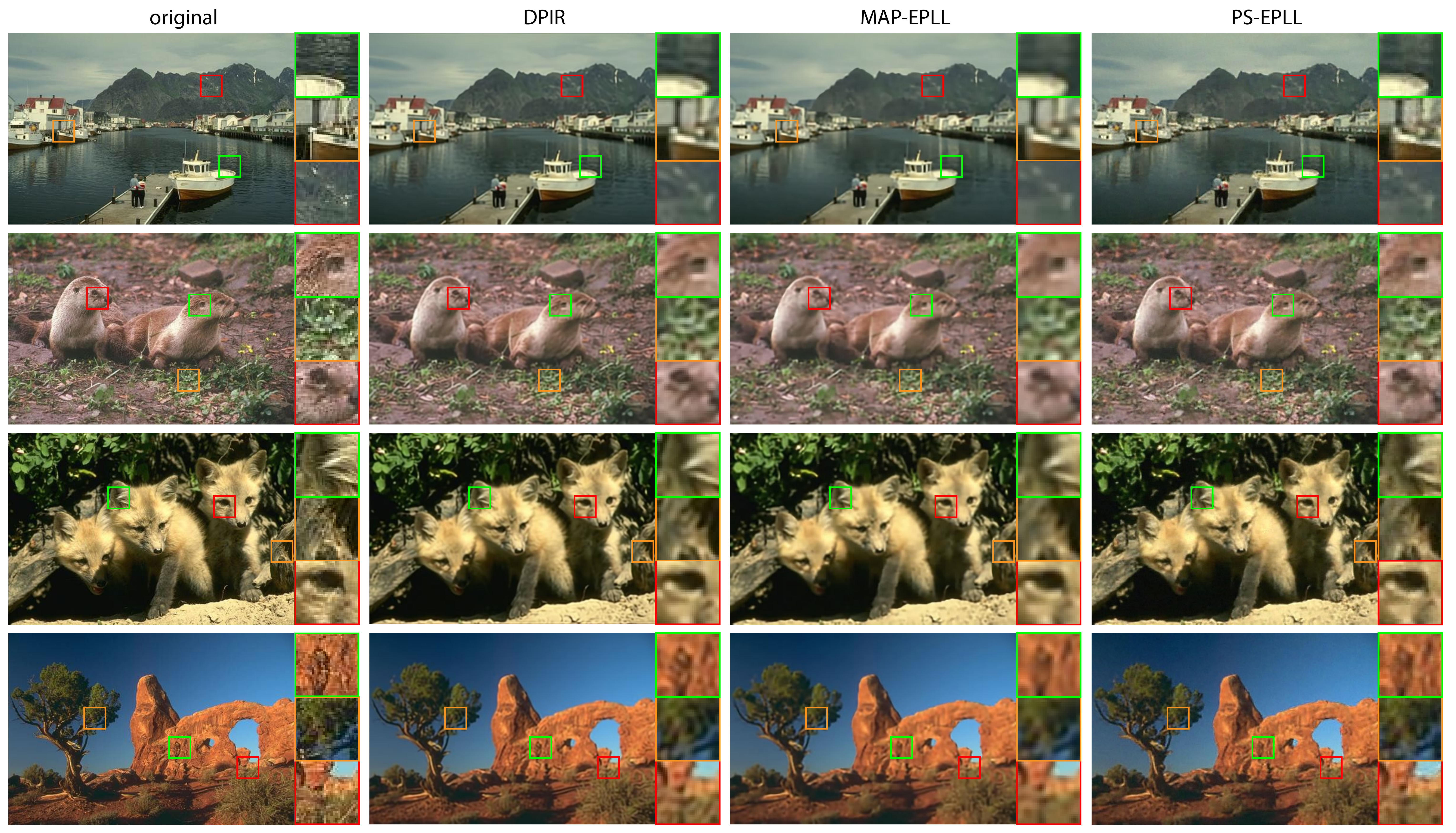}
\end{center}
    \caption{More examples of image deblurring for a Gaussian blur kernel with scale $s=2$}
    \label{fig:deblur-s2}
\end{figure*}

\begin{figure*}[h!]
\begin{center}
\includegraphics[width=\linewidth]{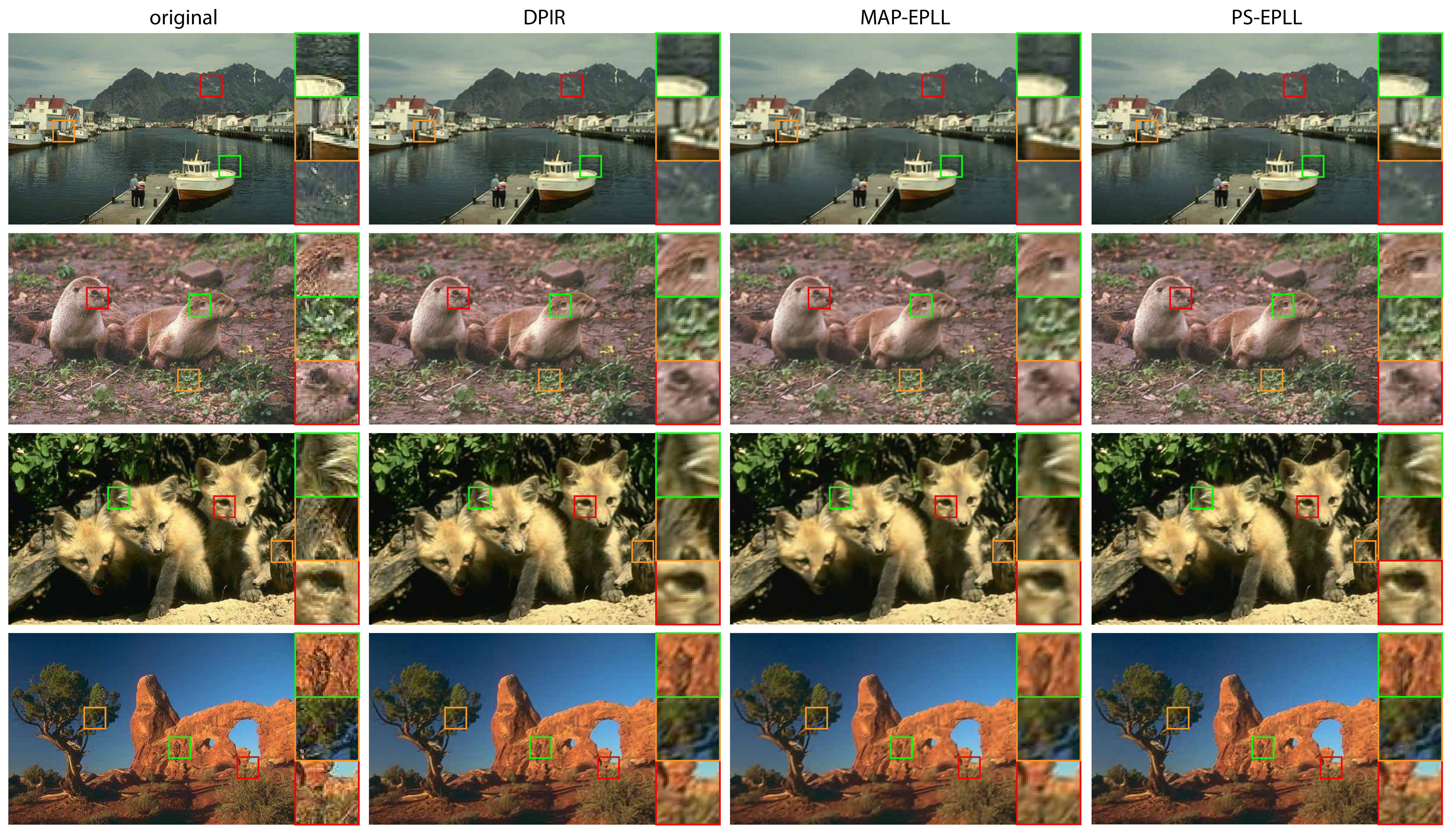}
\end{center}
    \caption{More examples of image deblurring for an elliptical Gaussian blur kernel}
    \label{fig:deblur-ellipse}
\end{figure*}

\begin{figure*}[h!]
\begin{center}
\includegraphics[width=\linewidth]{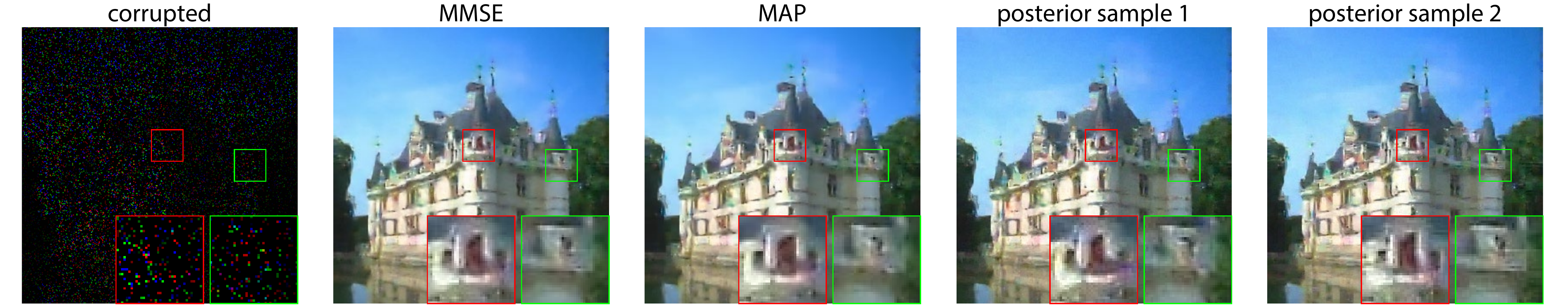}
\end{center}
    \caption{Example of inpainting an image with 95\% of its pixels missing.}
    \label{fig:inpainting}
\end{figure*}

\begin{figure*}[h!]
\begin{center}
\includegraphics[width=\linewidth]{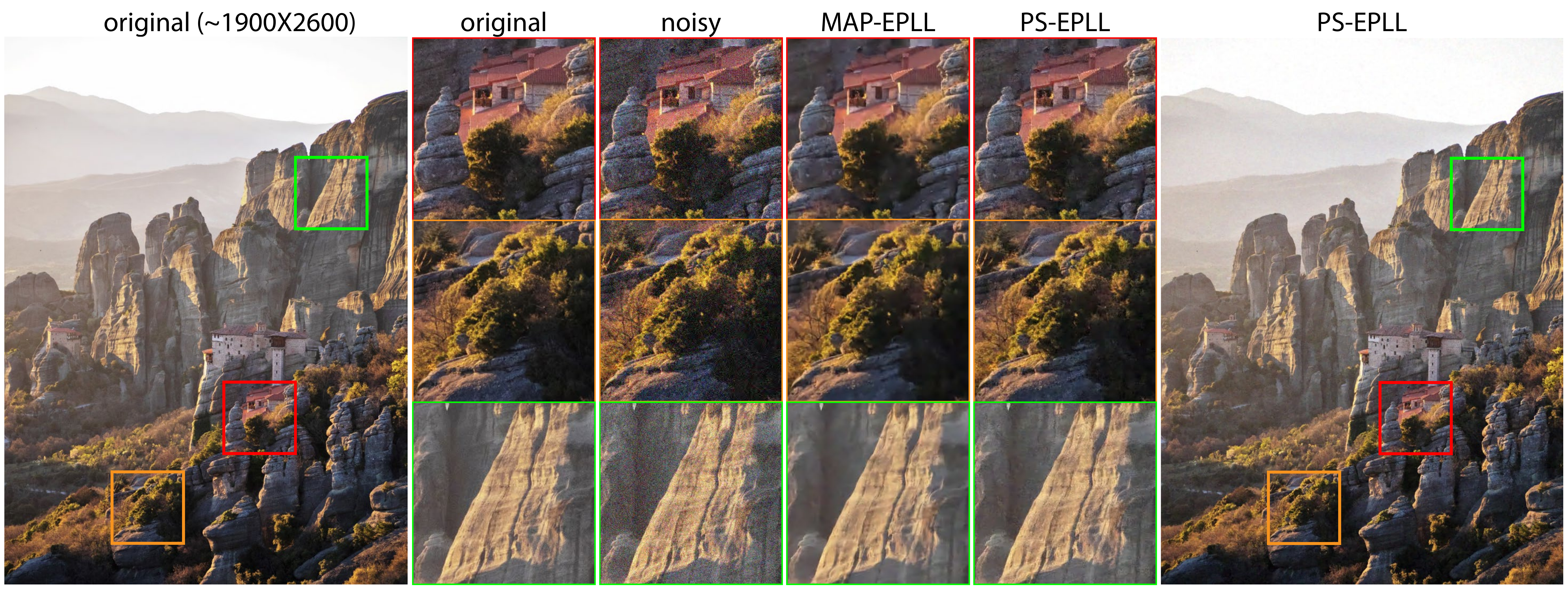}
\end{center}
    \caption{Another example of denoising a high resolution with $\sigma=25$ noise.}
    \label{fig:denoise-HR-n25}
\end{figure*}

\begin{figure*}[h!]
\begin{center}
\includegraphics[width=\linewidth]{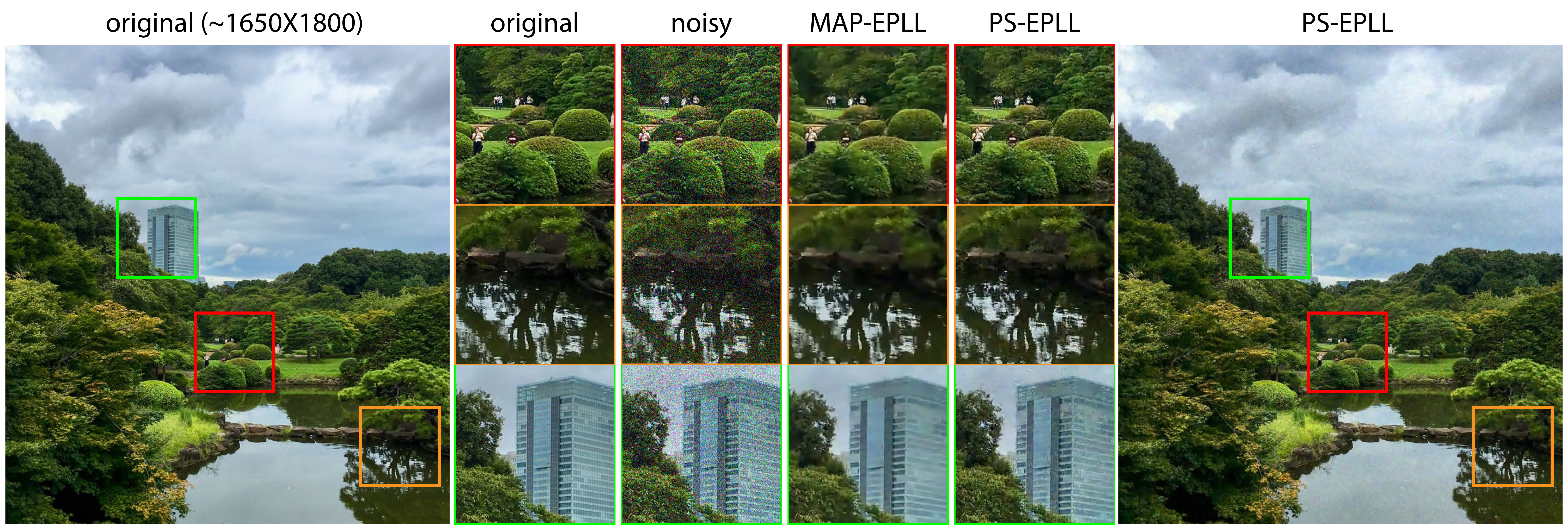}
\end{center}
    \caption{Another example of denoising a high resolution with $\sigma=50$ noise.}
    \label{fig:denoise-HR-n50}
\end{figure*}

Figures~\ref{fig:denoise-n15}-~\ref{fig:denoise-HR-n50} show more examples of image denoising, deblurring and inpainting using our approach.

\end{appendices}

\end{document}